\pdfoutput=1
\documentclass[format=sigconf, review=false, anonymous=false]{acmart}

\usepackage{tikz}
\usepackage{subcaption}
\usepackage{textcomp}
\AtBeginDocument{%
  \providecommand\BibTeX{{%
    \normalfont B\kern-0.5em{\scshape i\kern-0.25em b}\kern-0.8em\TeX}}}

\usepackage[suppress]{color-edits}
\addauthor[Amanda]{am}{blue}
\addauthor[Alex]{ac}{purple}
\addauthor[Edward]{ek}{green}
\addauthor[Alan]{al}{red}

\newtheorem{theorem}{Theorem}
\newcommand{\p}{\mathbb{P}}
\newcommand{\e}{\mathbb{E}}
\newcommand{\yh}{\hat{Y}}

\theoremstyle{condition}
\newtheorem*{condition}{Condition}

 \copyrightyear{2020}
\acmYear{2020}
\setcopyright{acmcopyright}
\acmConference[FAT* '20]{Conference on Fairness, Accountability, and Transparency}{January 27--30, 2020}{Barcelona, Spain}
\acmBooktitle{Conference on Fairness, Accountability, and Transparency (FAT* '20), January 27--30, 2020, Barcelona, Spain}
\acmPrice{15.00}
\acmDOI{10.1145/3351095.3372851}
\acmISBN{978-1-4503-6936-7/20/02}
 
\begin{document}

\title{Counterfactual Risk Assessments, Evaluation, and Fairness}

\author{Amanda Coston}
\email{acoston@cs.cmu.edu}
\affiliation{%
  \institution{  Heinz College and Machine Learning Department \\
  Carnegie Mellon University}
}

\author{Alan Mishler}
\affiliation{%
  \institution{Department of Statistics \\
  Carnegie Mellon University}
}

\author{Edward H. Kennedy}
\affiliation{%
  \institution{Department of Statistics \\
  Carnegie Mellon University}
}

\author{Alexandra Chouldechova}
\affiliation{%
  \institution{Heinz College \\
  Carnegie Mellon University}
}
\begin{abstract}
  Algorithmic risk assessments are increasingly used to help humans make decisions in high-stakes settings, such as medicine, criminal justice and education.  In each of these cases, the purpose of the risk assessment tool is to inform actions, such as medical treatments or release conditions, often with the aim of reducing the likelihood of an adverse event such as hospital readmission or recidivism.  Problematically, most tools are 
  trained and evaluated on historical data in which the outcomes observed depend on the historical decision-making policy.  These tools thus reflect risk under the historical policy, rather than under the different decision options that the tool is intended to inform.   Even when tools are constructed to predict risk under a specific decision, they are often improperly evaluated as predictors of the target outcome. 
  
  Focusing on the evaluation task, in this paper we define counterfactual analogues of common predictive performance and algorithmic fairness metrics that we argue are better suited for the decision-making context.  We introduce a new method for estimating the proposed metrics using doubly robust estimation. 
  We provide theoretical results that show that only under strong conditions can fairness according to the standard metric and the counterfactual metric simultaneously hold.  Consequently, fairness-promoting methods that target parity in a standard fairness metric may---and as we show empirically, do---induce greater imbalance in the counterfactual analogue.  We provide empirical comparisons on both synthetic data and a real world child welfare dataset to demonstrate how the proposed method improves upon standard practice. 
  
  
\end{abstract}

\settopmatter{printfolios=true}
\maketitle

\section{Introduction}

Much of the activity in using machine learning to help address societal problems focuses on algorithmic decision-making and algorithmic decision support systems. In settings such as health, education, child welfare and criminal justice, decision support systems commonly take the form of risk assessment instruments (RAIs), which distill rich case information into risk scores that reflect the likelihood of the case resulting in one or more adverse outcomes.  \cite{chouldechova2018case, kube2019allocating, ferguson2016policing, kehl2017algorithms, stevenson2018assessing, caruana2015intelligible, smith2012predictive}.
Prior literature has raised significant concerns regarding the fairness, transparency, and effectiveness of existing RAIs \amdelete{algorithmic risk assessments} \cite{barocas2016big, barabas2017interventions, dressel2018accuracy, corbett2017algorithmic, chouldechova2018frontiers}. Yet RAIs remain very popular in practice, and there is a large body of research on fairness and transparency promoting methods that seek to address some of these concerns \cite[e.g][]{zafar2015fairness, hardt2016equality, kamiran2012data, pleiss2017fairness, kamishima2011fairness, zemel2013learning}.

This paper highlights a different issue, one that has not received sufficient attention in the discussion of RAIs
but that nonetheless has significant implications for fairness:  RAIs are typically trained and evaluated as though the task were prediction when in reality the associated decision-making tasks are often interventions.  Models trained and evaluated in this way answer the question: What is the likelihood of an adverse outcome \textit{under the observed historical decisions}?  Yet the question relevant to the decision maker is: What is the likelihood of an adverse outcome \textit{under the proposed decision}? When decisions do not impact outcomes---when we are in what \cite{kleinberg2015prediction} call a ``pure predition'' setting---these are one and the same.  However, many decisions take the form of interventions specifically designed to mitigate risk.
RAIs for these settings must \amdelete{therefore} be developed and evaluated taking into account the effect of historical decisions on the observed outcomes. Failure to do so will result in RAIs that, despite appearing to perform well according to standard evaluation practices, underperform on cases such as those that have been historically receptive to intervention. 

In this paper we propose an approach to \textbf{counterfactual risk modeling and evaluation} to properly account for these intervention effects. Counterfactual modeling has been proposed for medical RAIs \cite{schulam2017reliable, shalit2017estimating, alaa2017bayesian}, and prior work has used counterfactual evaluation for off-policy learning in bandit settings \cite{dudik2011doubly}. 
 However, the question of adapting counterfactual evaluation for risk assessments and in particular for predictive bias assessments remains open.
In this paper, we propose a new evaluation method for RAIs that uses doubly-robust estimation techniques from causal inference \cite{van2003unified, robins2001inference}.  
We also argue that fairness metrics that are functions of the outcome should be defined counterfactually, and we use our evaluation method to estimate these metrics. 
We theoretically and empirically characterize the relationship between the standard fairness metrics and their counterfactual analogues.
Our results suggest that in many cases, achieving parity in the standard metric will not achieve parity in the counterfactual metric.


Our main contributions are as follows: 1) We define counterfactual versions of standard predictive performance metrics and propose doubly-robust estimators of these metrics (\textsection~\ref{sec:models}); 2) We provide empirical support that this evaluation outperforms existing methods using a synthetic dataset and a real-world child welfare hotline screening dataset (\textsection~\ref{sec:models}); 3) We propose counterfactual formulations of three standard fairness metrics that are more appropriate for decision-making settings (\textsection~\ref{sec:fair}); 4) We provide theoretical results showing that only under strong conditions, which are unlikely to hold in general, does fairness according to standard metrics imply fairness according to counterfactual metrics (\textsection~\ref{sec:fair}); 5) We demonstrate empirically that applying existing fairness-corrective methods can increase disparity in the counterfactual redefinition of the metric they target (\textsection~\ref{sec:fair}). 

\section{Background and Related Work} \label{sec:lit}

\subsection{Counterfactual learning and evaluation}
Literature on contextual bandits has considered counterfactual learning and evaluation of decision \textit{policies}.  While this literature is methodologically relevant, as we discuss below, it addresses a different problem.  In the \textit{decision support} setting we are considering, human users will ultimately decide what action to take.  The goal of the learning and evaluation task is not to learn a decision policy, but rather to learn a risk model that will inform human decisions.  That is, the risk assessment task is to accurately and fairly estimate the probability of an outcome under a given intervention.
 
 While the underlying task is different, the statistical methods used in evaluation are related. \cite{swaminathan2015batch} use propensity score weighting, a form of importance sampling, to correct for the effect of the historical treatment on the observed outcome, and they propose learning the optimal policy based on the minimization of the propensity-score weighted empirical risk. 
 Propensity-score methods are a good candidate when one has a good model of the historical decision-making policy, but may otherwise be biased. Doubly robust (DR) methods, by contrast, are robust to parametric misspecification of the propensity score model if instead one has the correct specification of the model of the regression outcome $\mathbb{E}[Y|X]$ where $Y$ is the outcome and $X$ are the features/covariates \cite{van2003unified, robins1994estimation, robins1995semiparametric}.  
 In a non-parametric setting, DR methods have faster rates of convergence than propensity-score methods \cite{kennedy2016semiparametric}. DR methods have been used for policy learning in the offline bandit setting \cite{dudik2011doubly}. The policy learned minimizes a DR estimate of the loss. Their framework can also be used to evaluate a policy by computing the DR estimate of its expected reward.
 

Prior work has considered counterfactual RAIs in   a temporal setting  \cite{schulam2017reliable}. 
In this work, the trained model is evaluated on real data using the observed outcomes, and on simulated data. 
Evaluating against the observed outcomes can be misleading in settings in which treatment was not assigned randomly (see \textsection~\ref{sec:count_eval}). 
In our work we propose instead to adapt DR techniques, as have been used in the bandit literature for evaluating policies, to provide evaluations of counterfactual RAIs.  

Counterfactual learning in the causal inference literature uses model selection based on DR estimation of counterfactual loss \cite{van2003unified}. Whereas this approach evaluates counterfactual metrics implicitly, our approach does so explicitly, providing the estimators for standard classification metrics in \textsection~\ref{sec:count_eval}.

There is also a line of work focused on counterfactual learning in the presence of hidden confounders. \cite{kallus2018confounding} propose policy learning via minimax regret learning over uncertainty sets. Their method is not immediately applicable to decision-support settings where RAIs are more informative to decision-makers than a policy recommendation. \cite{madras2019fairness} propose using deep latent variable models to model hidden confounders via proxies in the data and evaluate how well this model learns an optimal policy. While their model may be used for learning a risk assessment model, they do not address how to evaluate the model in such a setting, which is the focus of our work. \textsection~\ref{sec:models} of our paper assumes no hidden confounders, and future work could attempt to incorporate these techniques for handling hidden confounders. We note that our theoretical analysis in \textsection~\ref{sec:fair} holds even in the presence of hidden confounding. 

\subsection{Fairness and causality}

A growing literature on counterfactual fairness has offered notions of fairness based on the counterfactual of the protected attribute (or its proxy) \cite{kusner2017counterfactual, wang2019equal, kilbertus2017avoiding}. In this work, a policy is considered fair if it would have made the same decision had the individual had a different value of the protected attribute (and hence, potentially different values of features affected by the attribute). In this setting, the treatment decision is the outcome, and the protected attribute is the `treatment'.  By contrast, we consider counterfactual treatment decisions and consider a future observation to be the outcome.\footnote{This distinction is also made in a survey of fairness literature \cite{mitchell2018prediction}.} 

 Another line of work considers unfair causal pathways between the protected attribute (or its proxy) and the outcome variable or target of prediction \cite{nabi2018fair, zhang2018fairness}. These papers characterize or explain discrimination via path-specific effects, which are defined by interventions on the protected attribute. We do not consider interventions on (i.e. counterfactuals of) the protected attribute; rather, we propose methods that account for interventions on treatment decisions in training and evaluation. 
 
 Fairness definitions based on the counterfactual of the protected attribute are not widely used in RAI settings for two reasons: one technical and one practical. The technical challenge is that the assumptions required to estimate these counterfactual metrics prohibit the use of important features, such as prior history, or require full specification of the structural causal model (SCM) \cite{zhang2018equality, kusner2017counterfactual, kusner2019making} These requirements are too restrictive for our settings of interest where we have insufficient domain knowledge to construct the SCM and where we are unable to disregard important predictors like prior history. More significantly, the practical concern is that these definitions are ill-suited for risk assessment settings like child welfare screening. As we discuss in \textsection~\ref{sec:fair}, decisions made based on the counterfactual protected attribute may cause further harm to the protected groups. 
 

Our work bears conceptual similarity to the analysis of residual unfairness when there is selection bias in the training data that induces covariate shift at test time as discussed in \cite{kallus2018residual}. In settings where cases are systematically screened out from the training set, such as loan approvals in which we do not get to see whether someone who was denied a loan would have repaid, they find that applying fairness-corrective methods is insufficient to achieve parity. We consider a different but related setting in which we observe outcomes for all cases, but these outcomes are under different treatments. We propose fairness definitions that account for the effect of these treatments on the observed outcomes, and analyze the conditions under which existing methods can achieve this notion of counterfactual fairness.


\section{Counterfactual Modeling and Evaluation} \label{sec:models}
Before proceeding to introduce the learning approaches and evaluation methods considered in this work, we pause to clarify the types of risk-based decision policies to which our evaluation strategy as presented is tailored, and provide some background on algorithm-assisted decision making in child welfare hotline screening.

RAIs typically inform human decisions either by identifying cases that are the most (or least) \textit{risky}, or by identifying cases that are the most (or least) \textit{responsive}.  The evaluation metrics we consider are most directly relevant in the paradigm where human decision-makers wish to intervene on the \textit{riskiest} cases.  However,  our method can readily be adapted (as discussed in \textsection~\ref{sec:eval}) for paradigms in which interventions are being targeted based on responsiveness.  

The motivating application for our work is child welfare screening.
Child welfare service agencies across the nation field over 4.1 million child abuse and neglect calls each year \cite{usdhhs}. Call workers must decide whether to ``screen in'' a call, which refers to opening an investigation into the family. The child welfare system is responsible for responding to all cases where there is significant suspicion that the child is in present or impending danger.  
The standard of practice is therefore to identify the \emph{riskiest} cases.     Jurisdictions in California, Colorado, Oregon and Pennsylvania are all in various stages of developing and integrating RAIs into their call screening processes.  The RAIs are trained on historical data to predict adverse child welfare outcomes, such as re-referral to the hotline or out-of-home foster care placement \cite{chouldechova2018case}.  The decision to investigate a call can affect the likelihood of the target outcomes.


\subsection{Notation}
We use $Y \in \{0,1\}$ to denote the observed binary outcome, and for exposition we assume $Y=1$ is the unfavorable outcome.
$T \in \{0,1\}$ denotes the decision which for simplicity we take to be binary.  We note, however, that DR estimation methods can be used in any treatment setting, including for continuous treatments such as dosing \cite{vanderweele2013causal, kennedy2017non}.     Throughout the remainder of the paper we will use the term `decision' and `treatment' interchangeably to aid in the exposition.   In describing counterfactual learning and evaluation, we rely on the potential outcomes framework common in causal inference \cite{rubin2005causal, neyman1923applications, kennedy2013improved}.  In this framework, $Y^t$ denotes the outcome under treatment $t$.  For any given case we only get to observe $Y^0$ or $Y^1$, depending on whether the case was treated. 
We will take $T = 0$ to be the baseline treatment, the decision under which it is relevant to assess risk.   Most risk assessment settings have a natural baseline, which is often the decision to not intervene.  For instance, in education one might wish to assess the likelihood of poor outcomes if a student is not offered support; in child welfare it is natural to assess the risk of re-referral if the call is not investigated. \accomment{Reviewers might be concerned that the framework only permits considerations of a binary treatment vs. no treatment regime.  That's not the case though.  Maybe here we say that $T$ denotes treatment, which will be taken as binary for the purpose of our discussion.  But the results can generalized beyond the binary treatment setting.  Am I missing some kind of conceptual barrier to generalizing beyond binary treatment?   } \amcomment{updated}
We refer to the baseline treatment as \emph{control} and the not-baseline treatment as \emph{treatment}. 
$X \in \mathcal{X} \subseteq \mathbb{R}^d$ denotes the covariates (or features) which may include a protected or sensitive attribute $A \in \{0, 1\}$.   $\pi(X) = \mathbb{P}(T=1 \mid X)$ denotes the propensity score, whose estimate we denote by $\hat \pi(X)$.
In the child welfare setting, $X$ contains call details and historical information on all associated parties, $T$ is whether the case is screened-in for investigation, and $Y$ is whether the case is re-referred to the hotline in a six-month period. We use subscripts $i$ to index our data; e.g., $X_i$ are the features for case $i$.  We use $\hat{Y} :  \mathcal{X} \mapsto \{0,1\}$ to denote our predicted label and $\hat{s}: \mathcal{X} \mapsto [0,1]$ to denote the predicted score which is the model's estimate of the target outcome (our RAI).\footnote{$\yh(X)$ is typically obtained by thresholding $\hat s(X)$.} 
\accomment{Readers may find it helpful to have examples of the quantities in a context.  E.g., $Y$ may be re-referral, $T$ is whether a case is investigated, $X$ is historical information on a case, etc.  }\amcomment{updated}
 
\subsection{Learning models of risk}
\label{sec:learn}
In this section we introduce ``observational'' (standard practice) and ``counterfactual'' forms of model training.  

\subsubsection{Observational}
The \textit{observational} RAI produces risk estimates by regressing $Y$ on $X$ for the entire observed dataset.  i.e., this RAI estimates $\mathbb{E}[Y \mid X]$.   
This model answers the question: What is the likelihood of an adverse outcome under the \textit{observed historical decisions}?  The observational RAI is ill-suited for guiding future decisions; it will, for instance, underestimate (baseline) risk for cases that were historically responsive to treatment.  
\accomment{So, $T$ could be one of the $X$'s in this type of analysis, which would in some way ``account'' for the effect of treatment on the outcome. If we wanted to be more precise, I think the issue is that this framing of risk just isn't coherent for the decision support context.  The observational model is responsive to the question: Which cases historically had the greatest risk?  But what we care about in informing future decision making is: Which cases would have the greatest risk if we don't step in and intervene?  (Or, for which cases can we produce the greatest reduction in risk through intervention?)  }\amcomment{updated}

\subsubsection{Counterfactual} \label{sec:count_learning}
The counterfactual model of risk estimates the outcome under the baseline treatment.  Our counterfactual model of risk targets $\mathbb{E}[Y^0 \mid X]$.  Even though we  only observe $Y^0$ or $Y^1$ for any given observation, we may nevertheless draw valid inference about both potential outcomes under a set of standard identifying assumptions\footnote{Identification is the process of using a set of assumptions to write a counterfactual quantity in terms of observable quantities}. These assumptions hold by design in our synthetic dataset, and we discuss why they may be reasonable in the child welfare setting under each point.

\begin{enumerate}
    \item Consistency: $Y = TY^1 + (1-T) Y^0$. \\
    This assumes there is no interference between treated and control units. This is a reasonable assumption in the child welfare setting since  opening an investigation into one case will not likely affect another case's observed outcome.\footnote{We set the treatment to be the same value for all children in a family.}
    
    \item Exchangeability: $Y^0 \perp T \mid X$. This assumes that we measured all variables $X$ that jointly influence the intervention decision $T$ and the potential outcome $Y^0$. This is an untestable assumption but it may be reasonable in the child welfare setting where the measured variables capture most of the information the call screeners use to make their decision (see  Section \ref{sec:cw} for more details).
    
    \item Weak positivity requirement: $\mathbb{P}(\pi(X) <1) = 1 $ requires that each example have some non-zero chance of the baseline treatment. 
    This can hold by construction in decision support settings. 
    We can filter out cases that violate this assumption since the decision for these cases is nearly certain.\footnote{
    Risk assessments are unnecessary for these cases since the decision-maker already knows what to do.}
\end{enumerate}

Our assumptions identify the target $\mathbb{E}[Y^0|X]= \mathbb{E}[Y|X, T=0]$.
\amdelete{
\begin{equation}
    \mathbb{E}[Y|X, T=0]
\end{equation}}

The counterfactual model estimates $\e[Y^0 \mid X]$ by computing an estimate of $\e[Y \mid X, T =0]$. We can train such a model by applying any probabilistic classifier to the control population. Since the control population may have a different covariate distribution than the full population, reweighing can be used to correct this covariate shift \cite{quionero2009dataset}. This may be useful in a setting with limited data or where model misspecification is a concern \cite{sugiyama2007covariate}.

\subsection{Evaluation} \label{sec:eval}
To evaluate how well our models of risk might inform decision-making in the paradigm where interventions should be targeted at the riskiest cases, we assess performance metrics such as precision, true positive rate (TPR), false positive rate (FPR), and calibration.\footnote{In the paradigm where interventions are to be targeted at the \textit{most responsive} cases, performance metrics such as discounted cumulative gain (DCG) or Spearman's rank correlation coefficients are more natural choices for evaluation.  DR estimates can be constructed for these metrics as well.  }
Since the task is to evaluate how well the model predicts risk under a baseline intervention, we specify the performance metrics in terms of $Y^0$. The target counterfactual TPR is
\begin{equation} \label{eqn:tpr_target}
    \e[\yh \mid Y^0 = 1]
\end{equation} 
The target counterfactual precision is 
\begin{equation}
    \e[Y^0 \mid \yh =1]
\end{equation}
 The target counterfactual FPR is 
 \begin{equation}
     \e[\yh \mid Y^0 = 0]
 \end{equation}  A model is well-calibrated in the counterfactual sense when \begin{equation} \label{eqn:calib_target}
     \mathbb{E}\big[Y^0 \mid r_1 \leq \hat{s}(X) \leq r_2\big] \approx \frac{r_1 + r_2}{2}
 \end{equation} where $r_1, r_2$ define a bin of predictions.
We describe two standard practice approaches for evaluation, noting why these approaches do not adequately estimate the counterfactual targets. We introduce our proposed approach that uses doubly robust (DR) estimation.\footnote{All evaluations are computed on a test partition that is separate from the train partition}
\subsubsection{Observational Evaluation} \label{sec:obs_eval}
\accomment{A fairness audience might prefer to have Recall called the TPR.  E.g., a precision-recall curve would plot precision vs the TPR.} \amcomment{updated}
A standard practice approach evaluates the model against the observed outcomes. 
An observational Precision-Recall (PR) curve plots observational precision, ${\mathbb{E}[Y \mid \hat{Y} = 1]}$, against observational TPR\footnote{TPR and recall are equivalent.}, $\mathbb{E}[\hat{Y} \mid Y = 1]$.
An observational ROC curve plots observational TPR against observational FPR ${\mathbb{E}[\hat{Y} \mid Y = 0]}$. An observational calibration curve plots ${\mathbb{E}[Y \mid r_1 < \hat{s}(X) < r_2]}$, the observational outcome rate for scores in the interval $[r_1, r_2]$.
The observational evaluation answers the question: Does the RAI accurately predict the likelihood of an adverse outcome under the \textit{observed historical decisions}?   This evaluation approach can be misleading since $Y \neq Y^0$.  For instance, it will conclude that a valid counterfactual model of risk under baseline performs poorly because its predictions will be systematically inaccurate for cases that are responsive to treatment.  

\subsubsection{Evaluation on the Control Population} \label{sec:control_eval} 
\accomment{If we call this ``Evaluation on the Control Population'' and then describe it as a form of counterfactual evaluation rather than ``counterfactual evaluations via control population'', does that mess things up?  It seems like the Figures all call it ``control'' so those won't need to be regenerated.} \amcomment{updated}
The standard practice counterfactual approach to evaluation
computes error metrics on the control population
\cite{schulam2017reliable}. 
The PR curve evaluated on the control population plots $\mathbb{E}[Y \mid \hat{Y} = 1, T = 0]$ against  $\mathbb{E}[\hat{Y} \mid Y = 1, T = 0]$, and the ROC and calibration curves are similarly defined by conditioning on $T=0$. When the control population is not representative of the full population (i.e. $T \not{\perp} X$), as is the case in non-experimental settings, this evaluation may be misleading since $\e[Y \mid T = 0] = \e[Y^0 \mid T = 0] \neq \e[Y^0]$. A method that performs well on the control population may perform poorly on the treated population (or vice-versa). In child welfare, cases where the perpetrator has a history of abuse are more likely to be screened in. Since there is more information associated with these cases, a model may be able to discriminate risk better for these cases than on cases in the control population with little history. \amcomment{is this a good example or do we want something in the other direction? I think either direction is compelling}

 \subsubsection{Doubly-robust (DR) Counterfactual Evaluation} \label{sec:count_eval}
 We propose to improve upon the control population evaluation procedure by using DR estimation to perform counterfactual evaluation using both treated and control cases.  This ensures that performance is assessed on a representative sample of the population. Our method estimates the counterfactual outcome for all cases and evaluates metrics on this estimate. Other approaches
 such as inverse-probability weighing (IPW) or plug-in estimates could be used for a counterfacutal evaluation, but DR techniques are preferable because they have faster rates of convergence for non-parametric methods, and for parametric methods they are robust to misspecification in one of the nuisance functions, which estimate treatment propensity $\pi(X)$ and the outcome regression $\e[Y^0 \mid X]$ \cite{robins1994estimation, robins1995semiparametric, kennedy2016semiparametric}. Under sample splitting and $n^{1/4}$ convergence in the nuisance function error terms, these estimates are $\sqrt{n}$-consistent and asymptotically normal. This enables us to compute confidence intervals (see \emph{Calibration} below for an example). \accomment{Can you provide some intuition to readers for where these formulas are coming from?  Currently they seem to descend from the heavens :)} \amcomment{updated}
 
We first consider estimates of the average outcome under control $\e[Y^0]$.
Under our causal assumptions in Section \ref{sec:count_learning}, $\e[Y^0] = \e[ \e[Y \mid X, T =0]]$. The plug-in estimate is: 
$$\frac{1}{n}\displaystyle \sum_{i=1}^n \hat{s}_0(X_i)$$
where $\hat{s}_0(X)$ denotes the score of our counterfactual model. The IPW estimate uses the observed outcome on the control population and reweighs the control population to resemble the full population: 
$$\frac{1}{n} \sum_{i=1}^n \frac{1-T_i}{1-\hat \pi(X_i)} Y_i$$
DR estimators\footnote{In survey inference, this is known as the generalized regression estimator~\cite{sarndal1989weighted}.} combine the plug-in estimate with an IPW-residual bias-correction term for the control cases:
\begin{equation} \label{dr_y0}
    DR_{Y^0} = \frac{1}{n} \sum_{i=1}^n \Big[ \frac{1-T_i}{1-\hat \pi(X_i)} (Y_i-\hat s_0(X_i)) + \hat s_0(X_i)\Big]
\end{equation}

Next we consider the counterfactual targets in Equations~\ref{eqn:tpr_target}-~\ref{eqn:calib_target}. We identify the target under our causal assumptions and then state the DR estimator. We emphasize the distinction that $\hat s$ is the score of any model we wish to evaluate whereas $\hat s_0$ is the score of our counterfactual model in \textsection~\ref{sec:count_learning}. 
\paragraph{TPR (Recall):}
 Counterfactual TPR is identified as
\begin{equation}
     {\mathbb{E}[\hat{Y} \mid Y^0 = 1]} = \frac{ \mathbb{E}\big[\hat{Y}\mathbb{E}[Y \mid X,  T = 0]\big]}{\mathbb{E} \Big[ \mathbb{E} [ Y \mid X, T = 0]\Big]} 
\end{equation}.

\amdelete{
The target counterfactual TPR is $${\mathbb{E}[\hat{Y} \mid Y^0 = 1]}$$ Using our causal assumptions, this is identified as 
\begin{equation}
     \frac{ \mathbb{E}\big[\hat{Y}\mathbb{E}[Y \mid X,  T = 0]\big]}{\mathbb{E} \Big[ \mathbb{E} [ Y \mid X, T = 0]\Big]} 
\end{equation}.}

The DR estimate for the numerator is 
\begin{equation}
    \frac{1}{n} \sum_{i=1}^n \hat{Y_i} \Big[\frac{1-T_i}{1-\hat{\pi}(X_i)}(Y_i - \hat{s}_0(X_i)) + \hat{s}_0(X_i) \Big] 
\end{equation}

The DR estimate for the denominator is $DR_{Y^0}$ in Equation~\ref{dr_y0}.

\paragraph{Precision:} \amdelete{The target counterfactual precision is $$ \mathbb{E}[Y^0 \mid \hat{Y} = 1] $$ Under our causal assumptions this is identified as $$\mathbb{E}[\mathbb{E}[Y \mid X, T = 0] \mid \hat{Y} =1 ]$$}
The target counterfactual precision is identified as
\begin{equation}
    \mathbb{E}[Y^0 \mid \hat{Y} = 1] = \mathbb{E}[\mathbb{E}[Y \mid X, T = 0] \mid \hat{Y} =1 ]
\end{equation}
\amcomment{is it helpful or confusing to have LHS and RHS?}

The DR estimator for precision is 
\begin{equation}
    \frac{\frac{1}{n}\sum_{i=1}^n \Big[\frac{1-T_i}{1-\hat{\pi}(X_i)} (Y_i-\hat{s}_0(X_i)) + \hat{s}_0(X_i) \Big]\mathbb{I}\{\hat Y_i =1\}}{ \p(\hat Y_i=1) }
\end{equation}
where $\mathbb{I}$ denotes the indicator function.
\paragraph{Calibration:} The target in Equation~\ref{eqn:calib_target} is identified as  $$ \mathbb{E} \big[ \mathbb{E} [Y \mid X, T= 0] \mid r_1 \leq \hat{s}(X) \leq r_2 \big]$$

The DR estimate for calibration is 

\begin{equation}
    \frac{\frac{1}{n} \sum_{i=1}^n \Big[\frac{1-T_i}{1-\hat{\pi}(X_i)} (Y_i-\hat{s}_0(X_i)) + \hat{s}_0(X_i) \Big] \mathbb{I}\{ r_1 \leq \hat{s}(X_i) \leq r_2\}}{\p( r_1 \leq \hat{s}(X_i) \leq r_2)}
\end{equation}

To compute the confidence interval for this estimate, we compute the number of data points in the bin $n_r = \sum_{i=1}^n \mathbb{I}\{ r_1 \leq \hat{s}(X_i) \leq r_2\}$ and the variance in the bin 
$$var(\phi_r) = var\Big(\frac{1-T_i}{1-\hat{\pi}(X_i)} (Y_i-\hat{s}_0(X_i)) + \hat{s}_0(X_i)\mid r_1 \leq \hat{s}(X_i) \leq r_2 \Big)$$.

Then we use the normal approximation to compute the interval:
$\pm z \sqrt{\frac{var(\phi_r)}{n_r}}$ where $z = 1.96$ for a 95\% confidence interval.

\paragraph{FPR:} \amdelete{The target counterfactual FPR is $${\mathbb{E}[\hat{Y} \mid Y^0 = 0]}$$ Under our causal assumptions, this is identified as 
\begin{equation}
     \frac{ \mathbb{E}\Big[\hat{Y} \mathbb{E}[1-Y \mid X,  T = 0]\Big]}{\mathbb{E} \Big[ \mathbb{E} [ 1-Y \mid X, T = 0]\Big]}  
\end{equation}}

The target counterfactual FPR is identified as
\begin{equation}
    \mathbb{E}[\hat{Y} \mid Y^0 = 0] = \frac{ \mathbb{E}\Big[\hat{Y} \mathbb{E}[1-Y \mid X,  T = 0]\Big]}{\mathbb{E} \Big[ \mathbb{E} [ 1-Y \mid X, T = 0]\Big]}  
\end{equation}

The DR estimator for the numerator is 
\begin{equation}
    \frac{1}{n} \sum_{i=1}^n  \hat{Y_i} \Big[\frac{1-T_i}{1-\hat{\pi}(X_i)}( \hat{s}_0(X_i)-Y_i) + (1 - \hat{s}_0(X_i)) \Big] 
\end{equation}

For the denominator we use $1-DR_{Y^0}$ where $DR_{Y^0}$ is in Eq~\ref{dr_y0}.

\subsection{Results} 
\amdelete{We present the results of these three evaluations on a synthetic dataset and our child welfare dataset. Comparing to the true counterfactual for the synthetic data, we find that our DR evaluation is more accurate than either the observational or control evaluations. For the experiments on the real world child welfare data, where we do not have access to all counterfactuals, we perform a comparison to expert assessment of risk to give further credence to the conclusions from our DR evaluation. }

\subsubsection{Synthetic example} \label{sec:synth_risk}
We begin with a synthetic dataset so that we can compare methods in a setting where we observe both potential outcomes.  We specify two groups with different treatment propensities, but the treatment is constructed to be equally effective at reducing the likelihood of adverse outcome ($Y = 1$) for both groups.  
We generate 100,000 data points $(X_i, Y_i^0, Y_i^1, T_i)$ where $X_i = (Z_i, A_i)$ and $Z_i \sim \mathcal{N}(0,1)$, a normal distribution with mean 0 and variance 1. $A_i \sim Bern(0.5)$, a Bernoulli with mean 0.5. ${Y^0_i \sim Bern(\sigma(Z_i-0.5))}$ where $\sigma(z) = \frac{1}{1+e^{-z}}$. ${Y^1_i \sim Bern(c\sigma(Z_i-0.5))}$ where $c=0.1$ controls the treatment effect. 
${T_i \sim Bern(\sigma(Z_i-0.5+ kA_i))}$ where $k=1.6$ describes the bias in treatment assignment toward group $A=1$.\footnote{We present results for alternative values of $c$ and $k$ in Appendix~\ref{sec:appendix_synthetic}. The offset $-0.5$ is to roughly balance the number of treated/control units} We set \mbox{$Y=TY^1 + (1-T)Y^0$}.
The base rates are \mbox{$\e[Y] = 0.17$}; \mbox{$\e[Y^0] = 0.4$}; and $\e[Y^1] = 0.04$. The treatment rates are \mbox{$\e[T] = 0.55$; $\e[T \mid a = 0] = 0.4$}; and \mbox{$\e[T \mid a = 1] = 0.71$}. 

We use logistic regression to train both the observational ${\e[Y \mid X]}$ and counterfactual models ${\e[Y^0 \mid X]}$ as well as the propensity model $\e[T \mid X]$. Under this choice of model, the propensity model and counterfactual model are both correctly specified, and accordingly, the plug-in and IPW estimates are both consistent in this setting. However, in practice, there is no way to know whether the models are correctly specified, so DR estimates are preferable for real-world settings.
We use $X = (Z,A)$ as the features.\footnote{In Appendix~\ref{sec:appendix_synth_treat} we include $T$ as a feature in the observational model to see if this can appropriately control for treatment effects, but we find that it does not.} 

\begin{figure*}
\centering
\begin{subfigure}{\textwidth}
  \centering
  \includegraphics[scale=0.5, trim={0cm 0.5cm 0cm 0cm},clip]{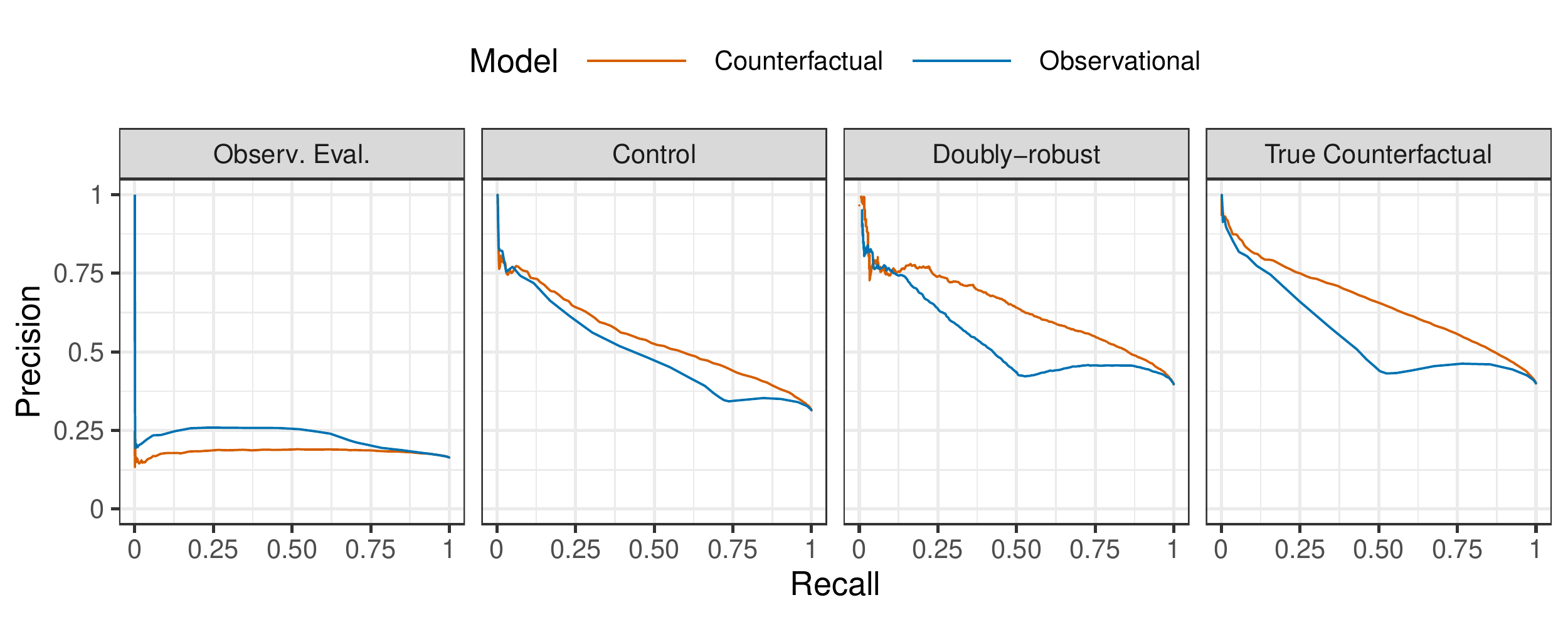}
    \caption{PR curves}
    \label{fig:synth_pr}
\end{subfigure}%
\\
\begin{subfigure}{\textwidth}
  \centering
  \includegraphics[scale=0.5, trim={0cm 0.6cm 0cm 0cm},clip]{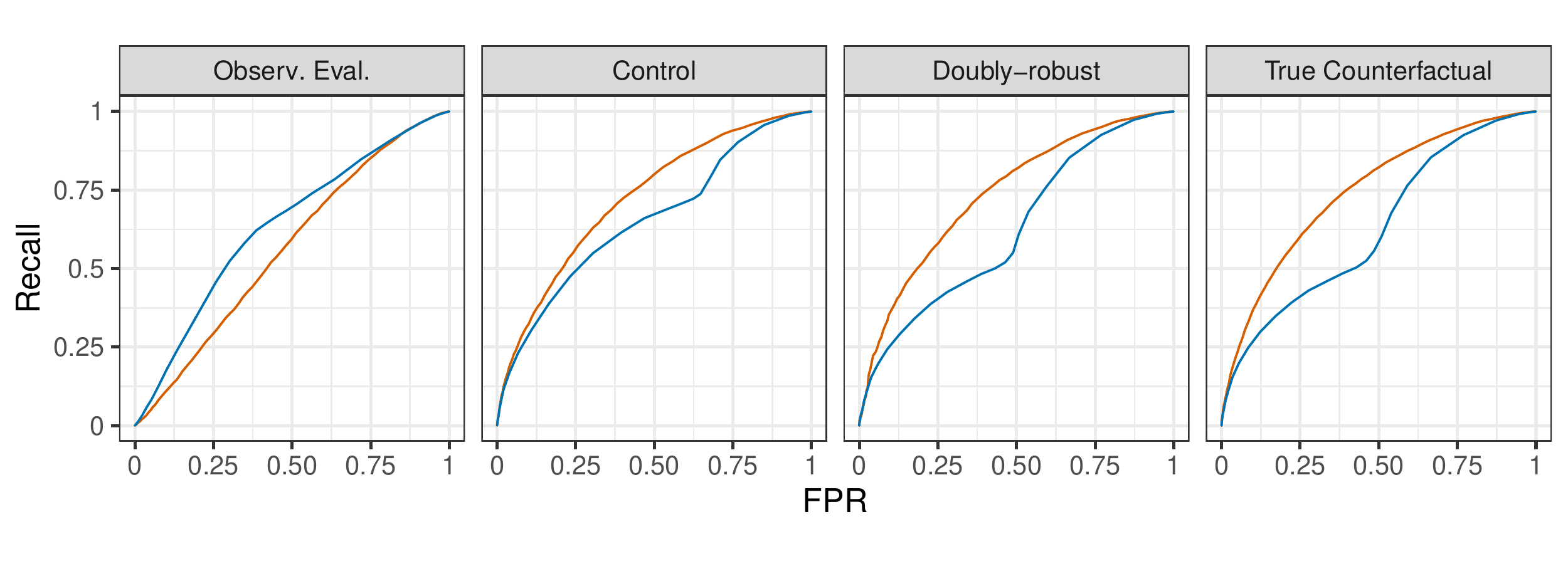}
  \caption{ROC curves}
  \label{fig:synth_roc}
\end{subfigure}
\\
\begin{subfigure}{\textwidth}
  \centering
  \includegraphics[scale=0.5, trim={0cm 0.6cm 0cm 0cm},clip]{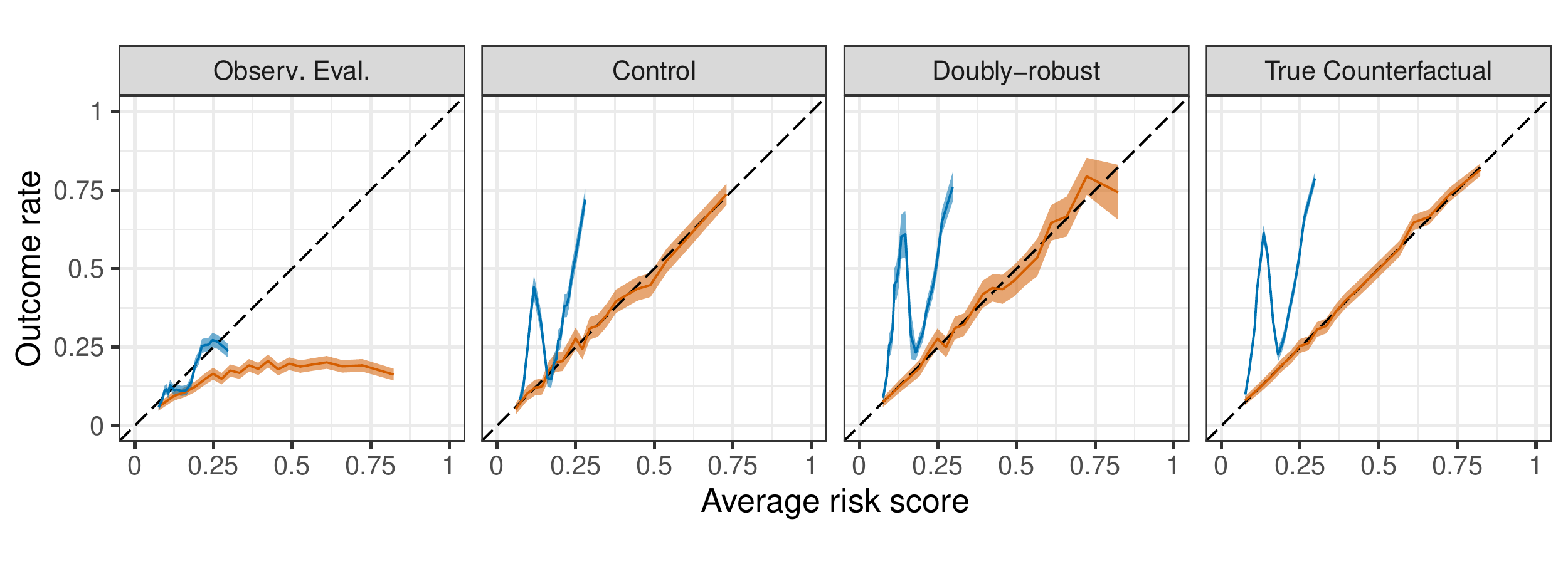}
  \caption{Calibration curves. 95\% pointwise confidence bounds shown.}
  \label{fig:synth_calib}
\end{subfigure}
\caption{Synthetic data results. Each column is an evaluation method (described in \textsection~\ref{sec:eval}). Colors denote the learning method (described in \textsection~\ref{sec:learn}). DR evaluation most accurately represents the true counterfactual evaluation. Observational evaluation erroneously suggests the observational model performs better than the counterfactual model because it evaluates against observed outcomes which includes units whose risk was mitigated by treatment. Control evaluation produces inaccurate curves because it does not assess how well the models perform on the treated population.  (See \textsection~\ref{sec:synth_risk} for details)}
\label{fig:synth}
\end{figure*}

Figure~\ref{fig:synth} displays PR, ROC, and calibration curves.\footnote{The code for this experiment is given in \href{https://github.com/mandycoston/counterfactual}{https://github.com/mandycoston/counterfactual}} DR evaluation most closely aligns with the true counterfactual evaluation. Notably, the observational evaluation suggests that the observational model outperforms the counterfactual model whereas the true counterfactual evaluation shows the counterfactual model performs better. 
 
\subsubsection{Child Welfare} \label{sec:cw}
We also apply counterfactual learning and evaluation to the problem of child welfare screening. 
The baseline intervention is screen-out (which means no investigation occurs).
The data consists of over 30,000 calls to the hotline in Allegheny County, Pennsylvania,  each containing more than 1000 features describing the call information as well as county records for all individuals associated with the call.
The call features are categorical variables describing the allegation types and worker-assessed risk and danger ratings. 
The county records include demographic information such as age, race and gender as well as criminal justice, child welfare, and behavioral health history. 
The outcome is re-referral within a six month period. Our approach contrasts to prior work which used placement out-of-home as the outcome \cite{chouldechova2018case, de2018learning}. 
This outcome is only observed for cases under investigation; therefore it cannot be used to identify $Y^0$, the risk under no investigation.

We use random forests to train the observational and counterfactual risk assessments as well as the propensity score model.  We used reweighing to correct for covariate shift but did not observe a boost in performance, likely because we have sufficient data and we used a non-parametric model.

\begin{figure*}
\centering
\begin{subfigure}{\textwidth}
  \centering
  \includegraphics[scale=.6, trim={0cm 0.3cm 0cm 0cm},clip]{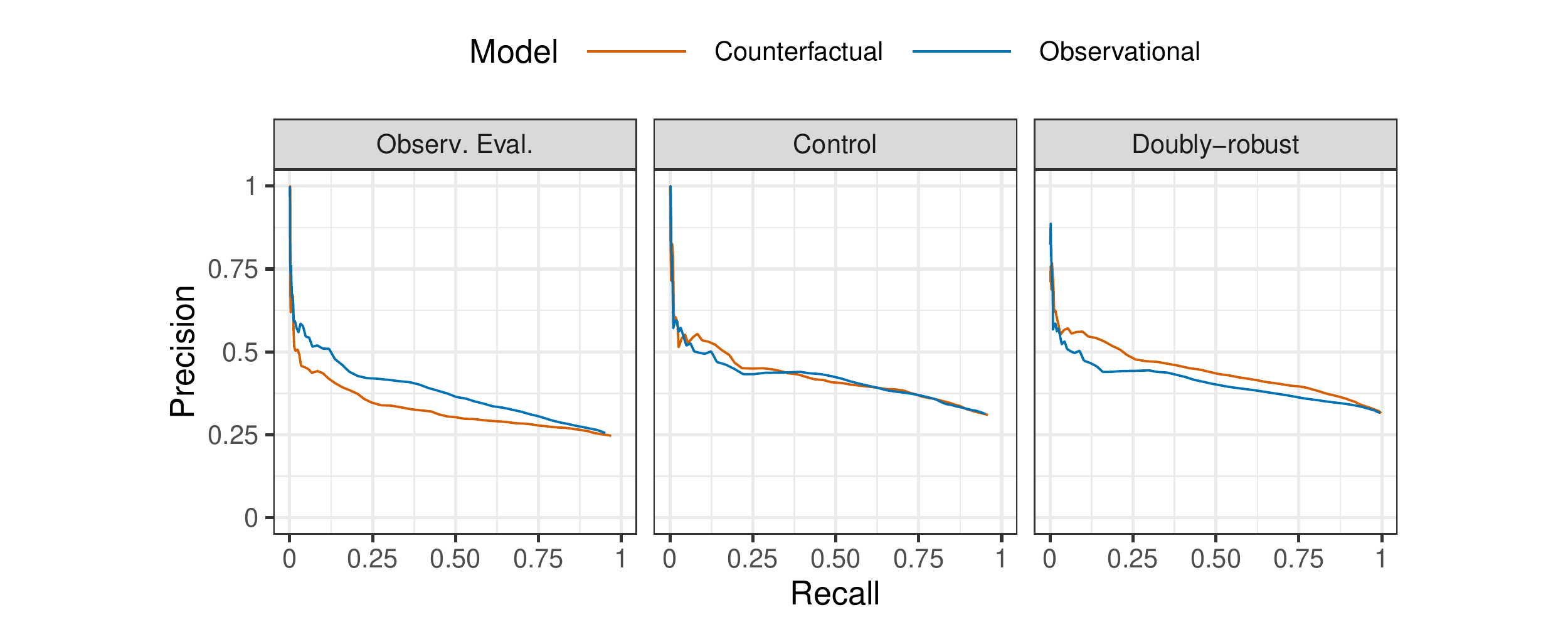}
    \caption{PR curves}
    \label{fig:cw_pr}
\end{subfigure}%
\\
\begin{subfigure}{\textwidth}
  \centering
  \includegraphics[scale=.55, trim={1cm 0.3cm 1cm 0cm},clip]{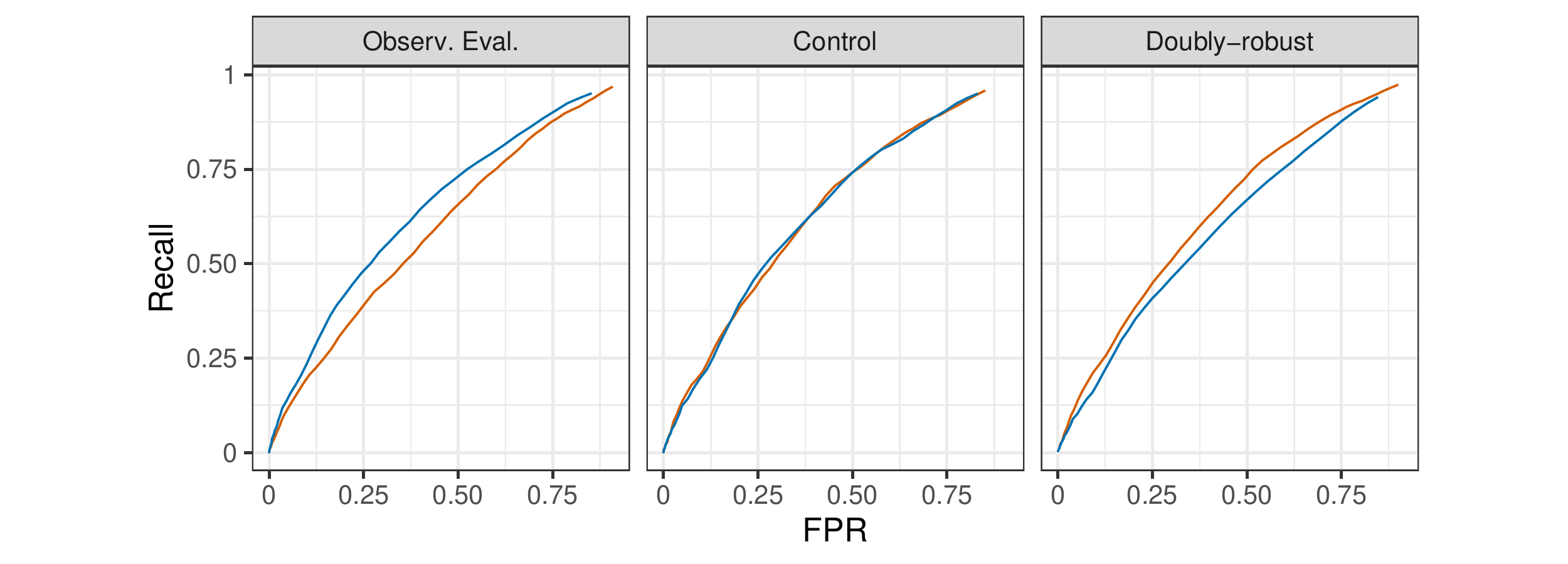}
    \caption{ROC curves}
    \label{fig:cw_roc}
\end{subfigure}
\\
\begin{subfigure}{\textwidth}
  \centering
  \includegraphics[scale=.45, trim={0cm 0.22cm 0cm 0cm},clip]{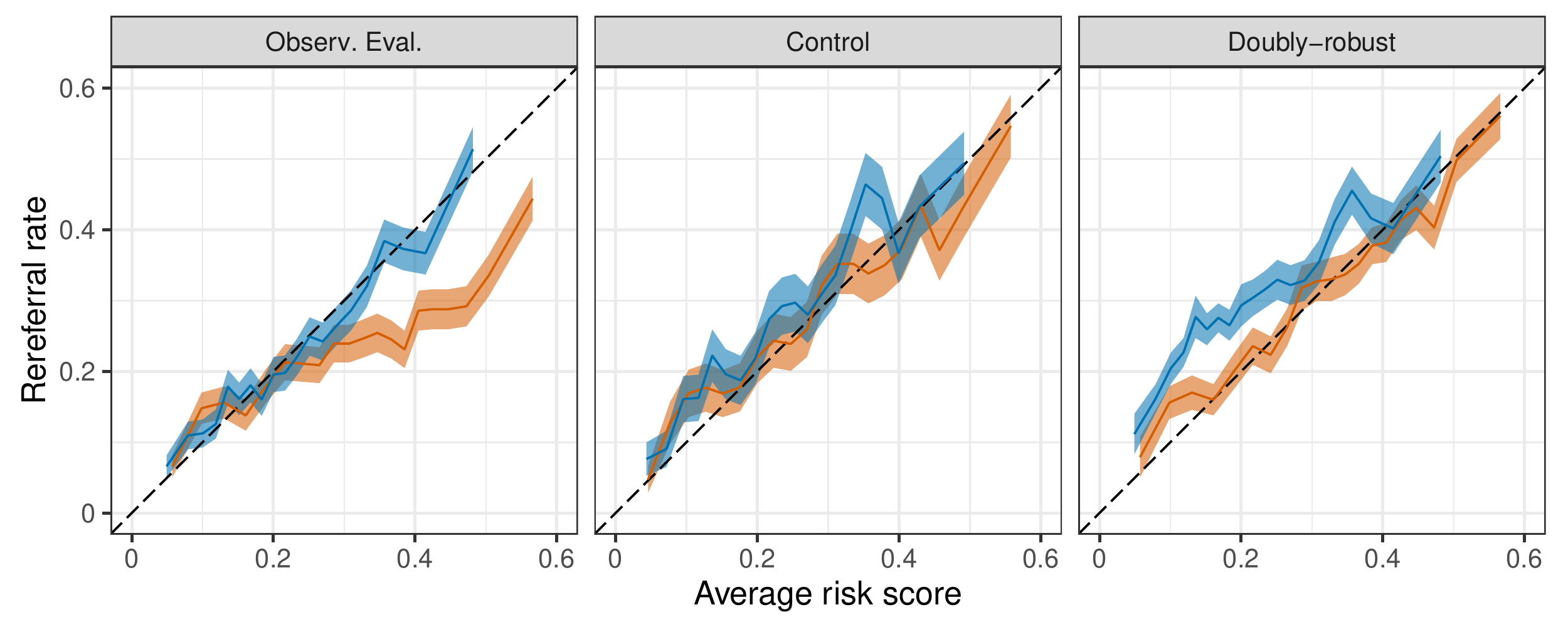}
    \caption{Calibration curves. 95\% pointwise confidence bounds shown.}
    \label{fig:cw_calib}
\end{subfigure}
\caption{Child welfare results. Each column is an evaluation method (\textsection~\ref{sec:eval}). Colors denote the learning method (\textsection~\ref{sec:learn}). Observational evaluation suggests the observational model has better discrimination and calibration than the counterfactual model because it evaluates against the observed outcomes which include cases whose risk was mitigated by child welfare services. Control evaluation suggests the two models perform similarly on cases that did not receive treatment. DR evaluation shows that the observational model does not perform well on treated cases. (See  Sections~\ref{sec:cw} and~\ref{sec:cw_expert} for details)}
\label{fig:cw}
\end{figure*}

We present the PR, ROC and calibration curves in Figure~\ref{fig:cw}.
The observational evaluation suggests that the observational model performs better. The control evaluation suggests that the counterfactual and observational models of risk perform equally well. 
Our DR evaluation suggests the counterfactual model has both better discrimination and calibration in estimating the probability of re-referral under screen-out.
In Figure~\ref{fig:cw_calib}, the observational evaluation suggests that the observational model is well-calibrated whereas the counterfactual model is overestimating risk; this is expected because the counterfactual model assesses risk under no investigation whereas the observed outcomes include cases whose risk was mitigated by child welfare services. The control evaluation suggests that the two models are similarly calibrated. 
The DR evaluation shows that the counterfactual model is well-calibrated and the observational model underestimates risk. This makes intuitive sense because the observational model is not accounting for that fact that treatment reduced risk for the screened-in cases.

We see further evidence that the observational model performs poorly on the treated population in the drop in ROC curves between the control evaluation and DR evaluation in Figure~\ref{fig:cw_roc}.
Deploying such a model would mean failing to identify the people who need and would benefit from treatment. 
The observational and control evaluations do not show this significant limitation; DR evaluation is the only evaluation that illustrates the poor performance of the observational model on the treated population. 

\begin{figure}
    \centering
    \includegraphics[scale=.6, trim={0.25cm 0.5cm 0cm 0.25cm},clip]{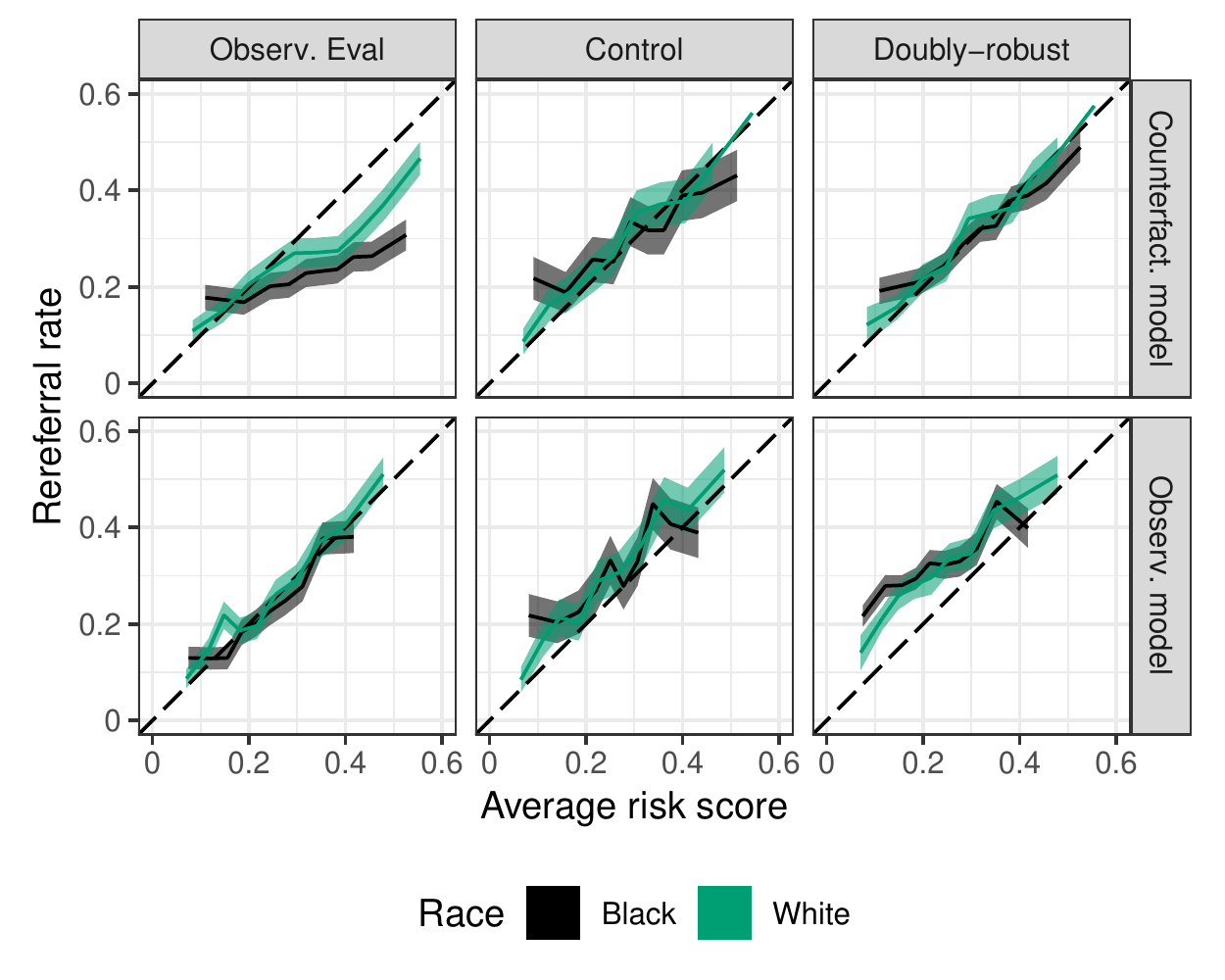}
    \caption{ Calibration curves by race for child welfare. Counterfactual model (top row) is well-calibrated by race according to the control and DR evaluations but shows inequities according to the observational evaluation because black cases were more likely to get treatment which mitigates risk (see \textsection~\ref{sec:cw} for more details). The observational model (bottom row) is poorly calibrated for both black and white cases according to the DR evaluation.}
    \label{fig:cw_calib_race_color}
\end{figure}

We also evaluate the different models according to whether they are equally predictive, in the sense of being equally well calibrated, across racial groups.  Research suggests child welfare processes may disproportionately involve black families \cite{dettlaff2011disentangling}. Here we ask whether the observational or counterfactual model is more equitable. We compare calibration rates by race in Figure ~\ref{fig:cw_calib_race_color}. The observational evaluation suggests that the counterfactual model of risk is poorly calibrated by race.
The DR evaluation shows that the counterfactual model is well-calibrated by race and indicates that the observational model underestimates risk on both black and white cases.

Overall the observational evaluation suggests that the observational model performs better whereas the DR evaluation suggests the counterfactual model performs better. Since we do not have access to the true counterfactual to validate these results, we further consider how well the models align with expert assessment of risk.


\subsubsection{Expert Evaluation} \label{sec:cw_expert}
\begin{figure}
    \centering
        \includegraphics[scale=.45, trim={0cm 0.5cm 0cm 0cm},clip]{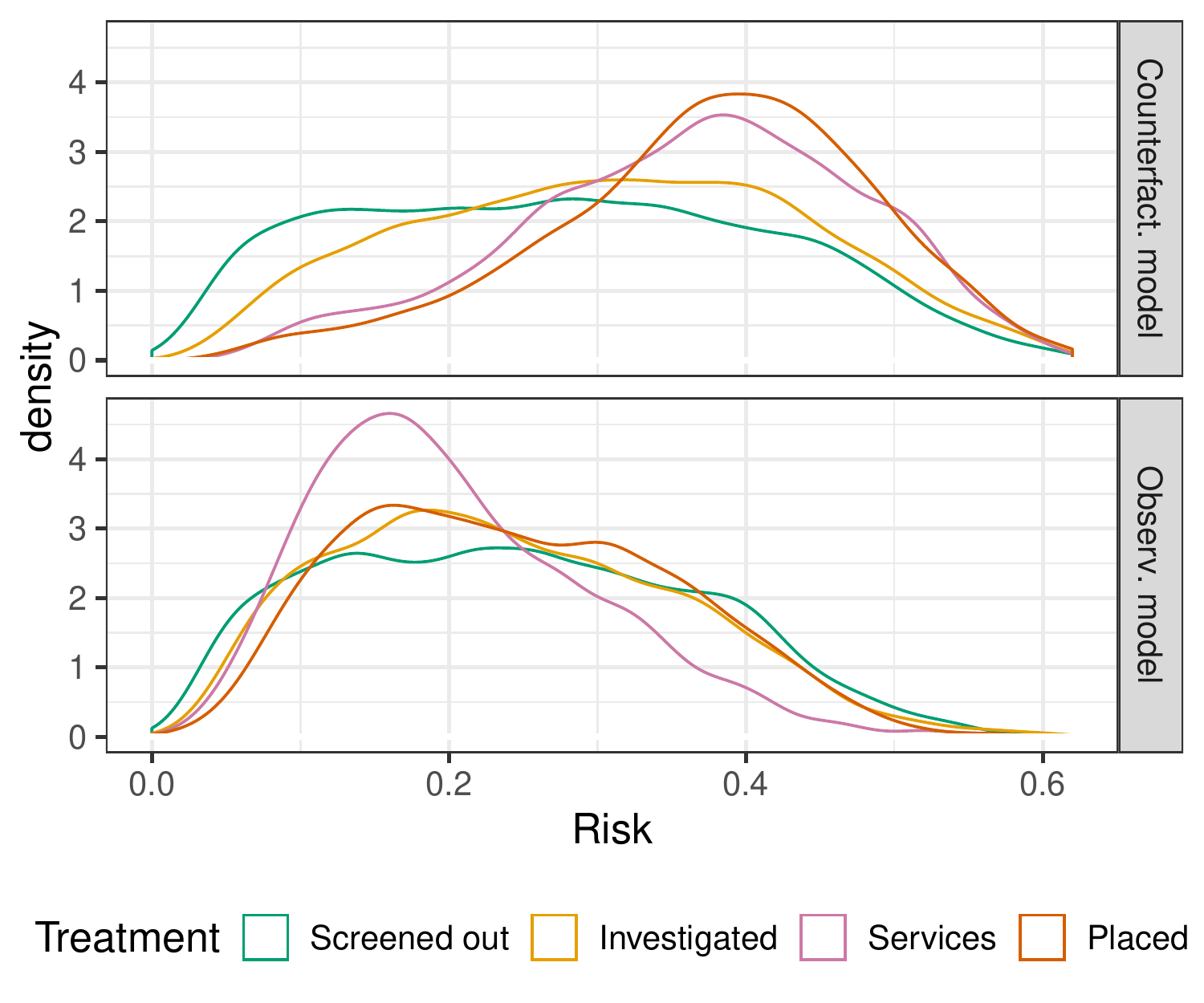}
    \caption{Child welfare risk distributions by treatment type for counterfactual and observational models. We expect risk to increase with the severity of treatment assigned, with `Placed' out-of-home having the highest risk distribution and `Screened out' of investigation having the lowest (see \textsection~\ref{sec:cw_expert}). The counterfactual model displays this expected trend whereas the observational model does not. The observational model underestimates risk on cases where child welfare effectively mitigated the risk}
    \label{fig:risk_dist}
\end{figure}

At various stages in the child welfare process, social workers assign treatment based on their assessment of risk. Social workers sequentially make three treatment decisions: 
\begin{enumerate}
    \item Whether to screen in a case for investigation
    \item Whether to offer services for a case under investigation
    \item Whether to place a child out-of-home after an investigation
\end{enumerate}
Assuming that social workers are competent at assessing risk, we expect the group placed out-of-home (3) to have the highest risk distribution, followed by the group offered services (2), followed by those screened in, and finally we expect the screened out group to have the lowest risk. Figure \ref{fig:risk_dist} shows that the counterfactual model exhibits this expected behavior whereas the observational model does not. 
 The observational model assesses the screened out population to have more high risk cases than any other treatment group.
This indicates that the observational model is underestimating risk on the treated groups (investigated, services, and placed) since it fails to account for the risk-mitigating effects of these treatments. 
The observational model underestimates risk on those who were assigned effective treatments. These cases \emph{should} be assigned treatment, but the observational model would suggest that they are low risk and should be screened out.

\begin{figure}
    \centering
    \includegraphics[scale=0.6, trim={0cm 0.6cm 0cm 0cm},clip]{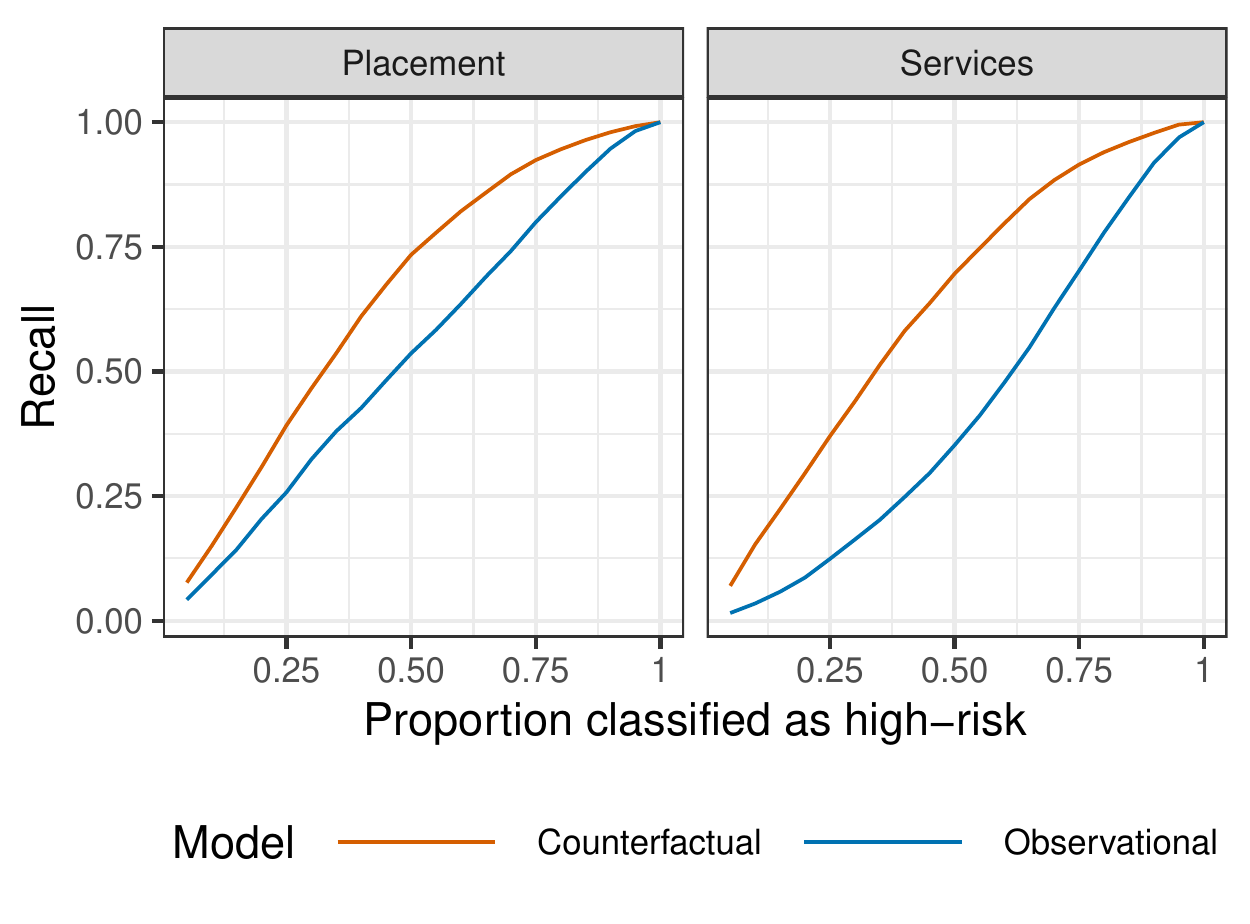}
    \caption{Recall for downstream child welfare decisions. At current screen-in rates (0.5), the observational model would screen out nearly 50\% of very high risk cases that were placed out-of-home. The counterfactual model has higher recall at 73\%. The gap is even larger for cases that were accepted for services. (See \textsection~\ref{sec:cw_expert}). }
    \label{fig:recall-downstream}
\end{figure}

Such a mistake can have cascading effects downstream.
We are particularly concerned about screening out cases that, had they been screened in, would have been accepted for services or placed out-of-home.
\amdelete{Humans determined that these cases needed treatments that will be inaccessible if they are screened out.}
Figure \ref{fig:recall-downstream} shows the recall for placed cases and serviced cases as we vary the proportion of cases classified as high-risk. This plot shows that at any proportion the counterfactual model has significantly higher recall for both services and placement cases. \amdelete{In particular, at the 0.5 proportion (which is the rate of screen in), the counterfactual model screens in 74\% of cases that were placed whereas the observational model only screens in 53\%. At the 0.5 proportion the counterfactual model screens in 69\% of cases that were accepted for services versus 31\% for the observational model.}

\subsubsection{Task adaptation: Predicting Placement} \label{sec:cw_task}
Another way to evaluate  the models is to assess their performance on related risk tasks.
While the counterfactual risk models $\mathbb{E}[Y^0|X]$, we can assess how well it estimates $\mathbb{E}[Y^1|X]$, which is the risk under investigation. 
If we have reason to believe there will be common risk factors for risk under no investigation and risk under investigation, then we expect our model to 
perform well on this
task. 
We use placement out-of-home, an adverse child welfare outcome that is observed for cases under investigation.

Table~\ref{fig:au-placement} shows the area under the ROC and PR curves for the placement task. The observational model performs worse than a random classifier, whereas the counterfactual model shows some degree of discrimination. 
This suggests that the counterfactual model is learning a risk model that is useful in related risk tasks whereas the observational model is not.

\begin{table}[ht]
\centering
\begin{tabular}{rllr}
  \hline
 & Observ. model & Counterfact. model & Random \\ 
  \hline
AUROC & 0.48 (0.46,0.49) & 0.62 (0.61,0.63) & 0.50 \\ 
  AUPR & 0.13 (0.11,0.14) & 0.18 (0.16,0.19) & 0.14 \\ 
   \hline
\end{tabular}
\caption{Area under ROC and PR curves using our re-referral models to predict a related risk task, out-of-home placement (95\% confidence intervals given in parentheses). The observational model performs worse than a random classifier. The counterfactual model performs better; it learns a model of risk that transfers to related risk tasks whereas the observational model does not.  (See \textsection~\ref{sec:cw_task})} 
\label{fig:au-placement}
\end{table}
The comparison to expert assessment of risk and the performance on a downstream risk task support the conclusions of our DR evaluation: the counterfactual model outperforms the observational model. In decision-making contexts, failure to account for treatment effects can lead one to the wrong conclusions about model performance, even potentially leading to the deployment of a model that underestimates risk for those who stand to gain most from treatment. In the next section, we consider how failure to account for treatment effects can impact fairness.

\section{Counterfactual Fairness} \label{sec:fair}
Standard observational notions of algorithmic fairness are subject to the same pitfalls as observational model evaluation. In this section we propose counterfactual formulations of several fairness metrics and analyze the conditions under which the standard (observational) metric implies the counterfactual one.  

We motivate the importance of defining these metrics counterfactually with an example.
 Suppose teachers are assessing the effectiveness and fairness of a model that predicts who is likely to fail an exam which they intend to use to assign tutoring resources. Suppose anyone tutored will pass. The tutoring session conflicts with girls' sports practice so only male students are tutored. A model that perfectly predicts who will fail without the help of a tutor will have a higher observational FPR for men than women because some male students were tutored, which enabled them to pass. It would be wrong to conclude that this model is unfair with regards to FPR. 
Someone who would have been high-risk had they not been treated but whose risk was mitigated under treatment should not be considered a false positive. Failure to make this distinction could lead to unfairness, not only in settings where the treatment assignment varies according to the protected attribute but also in settings where the risk under treatment varies according to the protected attribute, as we can see in the next example.

 Suppose that the classroom next door is also evaluating the model. This classroom offers tutoring during lunch so girls and boys both can attend; however they hired a tutor who happens to only be effective in preparing male students to pass. The teachers don't know this and randomly assign this tutor to students regardless of gender. The model that perfectly predicts who will fail without a tutor has a higher observational FPR for men, but as before, it is wrong to conclude that the model is unfair with regards to FPR. 
 
 \amdelete{These examples give intuition for the theory presented in the next Section. As we show, when there
 are differences in the propensity to treat
 and/or the treatment is differentially effective, parity in the observational metric generally implies counterfactual disparity. }
\amdelete{This suggests that methods that attempt to equalize the observational metric may not be equalizing the counterfactual metric and raises the question whether they could increase the disparity in the counterfactual metric. In Section ~\ref{sec:fair_experiments}, we perform experiments on a synthetic dataset that illustrate examples in which this occurs.}

We distinguish our notion of counterfactual fairness from prior work which considered counterfactuals of the protected attribute \cite{kusner2017counterfactual, kilbertus2017avoiding, wang2019equal}, an approach which is counterproductive in our settings of interest. Consider a female student who is at high risk of failing because of gender discrimination at home or in the classroom e.g. parents or previous teachers have not given her the support they would have had she been male. Treating this student "counterfactually as if she had been male all along" may suggest that we should not assign this student a tutor. In fact we \emph{must} assign her a tutor in order to correct historical discrimination. Similar arguments can be made in settings like child welfare screening and loan approvals. 
\subsection{Theoretical results} \label{sec:fair_theorems}
For three definitions of fairness (parity), we show that observational parity implies counterfactual parity if and only if a balance condition holds. We further show that an independence condition is sufficient for observational parity to imply counterfactual parity. We discuss why it is generally unlikely that the independence condition holds and even more unlikely that the finer balance condition holds when the independence condition fails. 
All proofs are provided in Appendix~\ref{appendix_proofs}. 

\subsubsection{Base Rate Parity} \label{sec:br_parity}
Base rate plays a core role in statistical definitions of fairness (also known as group fairness). 
Base rate parity is similar to the fairness notion of demographic parity, which requires $\yh \perp A$ \cite{dwork2012fairness, calders2009building, zafar2015fairness}. In Section ~\ref{sec:fair_experiments}, we perform experiments on a fairness corrective method that targets base rate parity in order to encourage demographic parity  \cite{kamiran2012data}.  A related fairness notion, prediction-prevalence parity, requires $\mathbb{E}[Y \mid a] = \mathbb{E}[\hat Y \mid a]$. Satisfying both prediction-prevalence parity and demographic parity requires parity in the base rates. \amcomment{Any further citations for these definitions, prediciton-prevalence in particular?} 
We distinguish observational base rate parity (oBP) $Y \perp A$ from  counterfactual base rate parity (cBP), which requires $Y^0 \perp A$, where $Y^0$ is the potential outcome under the baseline treatment. 
\begin{theorem}[Base Rate Parity] \label{thm:BP} Assume $\p(T=0 \mid y^0, a) \neq 0$. If oBP holds, then cBP holds if and only if the following balance condition holds.
\begin{condition}[balBP]
\begin{equation} \label{eq:balBP}
     \begin{split}
         &\p(Y^1 = y)\p(T=1 \mid Y^1 = y) - \p(Y^1 = y \mid a)\p(T=1 \mid Y^1 =y, a) \\&= \p(Y^0 = y)\Big( \p(T=1 \mid Y^0 = y) - \p(T=1 \mid Y^0 =y, a) \Big)  
     \end{split}
 \end{equation}
\end{condition}

BalBP holds under the following independence conditions, which provide sufficient conditions for oBP to imply cBP. 
\begin{condition}[indBP]
\begin{equation} \label{indBP}
\begin{split}
    & T \perp A \mid  Y^0 \\
    & (Y^1, T) \perp A
\end{split}
\end{equation}
\end{condition}
\end{theorem}
It is unlikely that indBP (\ref{indBP}) holds in many contexts. In settings such as child welfare screening and criminal justice, research suggests that even when controlling for the true risk, certain races are more likely to receive treatment \cite{dettlaff2011disentangling, alexander2011new, mauer2010justice}. indBP cannot hold in these settings since $T \not{\perp} A \mid  Y^0$. Even in settings where there is no 
such bias,
indBP will not hold if the risk distributions under treatment vary by protected attribute since indBP requires that $Y^1 \perp A$. indBP also requires $T \perp A \mid Y^1$, which forbids discrimination in treatment assignment when controlling for risk under treatment. If indBP does not hold, it is possible that balBP (\ref{eq:balBP}) still holds if the conditional and marginal probabilities are such that all terms in Condition \ref{eq:balBP} exactly cancel; however there is no semantic reason why this should hold. 
Theorem 1 assumes $\p(T=0 \mid y^0, a) \neq 0$, a mild positivity-like assumption that holds in all settings that are  suitable for algorithmic risk assessment. Violations of this assumption indicate either completely perfect or imperfect treatment assignment historically for a demographic group.

\amdelete{
\begin{proof} [Proof of Base Rate Necessary Condition]
 By consistency $Y = T Y^1 + (1-T) Y^0$. Then we have $\mathbb{P}(Y = y)=$ 
 \begin{equation*}\label{eqn:nBPo} \mathbb{P}(y^1)\mathbb{P}(T= 1 \mid y^1) +\mathbb{P}(y^0)\mathbb{P}(T= 0 \mid y^0) 
 \end{equation*}
 Likewise for $\mathbb{P}(Y = y \mid a) =$ 
 \begin{equation*} \label{eqn:nBPa}
    \begin{split}
    &\mathbb{P}(y^1 \mid a)\mathbb{P}(T= 1 \mid y^1, a) +\mathbb{P}(y^0 \mid a)\mathbb{P}(T= 0 \mid y^0, a) 
    \end{split}
 \end{equation*}
By oBP, $\mathbb{P}(Y = y)= \mathbb{P}(Y = y \mid a)$. We assume cBP holds so $\mathbb{P}(y^0) = \mathbb{P}(y^0 \mid a)$. Then, we have 
  \begin{equation*} \label{eq:BR}
     \begin{split}
         &\p(y^1)\p(T=1 \mid y^1) - \p(y^1 \mid a)\p(T=1 \mid y^1, a) \\&= \p(y^0)\Big( \p(T=1 \mid y^0) - \p(T=1 \mid y^0, a) \Big)
     \end{split}
 \end{equation*}
\end{proof}

\begin{proof}[Proof of Base Rate Parity Sufficiency]
 \begin{equation*}
     \begin{split}
        \p(Y = 1 \mid a ) &= \p(TY^1 + (1-T) Y^0 =1 \mid a) \\
     &= \p(TY^1 =1)  + \p\big((1-T) Y^0 =1\mid a\big) \\
     \end{split}
 \end{equation*}
 where the first line used consistency and the second line applied linearity of expectation and $ (Y^1, T) \perp A$.
 By oBP, $\p(Y=1) = \p(Y=1 \mid a)$, so it must be true that  
 \begin{equation*}
     \begin{split}
         &\p\big((1-T)Y^0 =1\big) = \p\big((1-T) Y^0 =1 \mid a\big)  \implies (T,Y^0 ) \perp A \\ &\implies Y^0 \perp A
     \end{split}
 \end{equation*}
 \end{proof}}

\subsubsection{Predictive parity}
Base parity and demographic parity may be ill-suited for settings where base rates differ by protected attribute due to disparate needs. Here we may instead desire parity in an error metric, such as precision. 
Positive predictive parity requires the precision (also known as positive predictive value) to be independent of the protected attribute, and negative predictive parity requires the negative predictive value to be independent of the protected attribute \cite{chouldechova2017fair, kleinberg2016inherent}. We define observational Predictive Parity (oPP) as $Y \perp A \mid \hat{Y} = \hat y $ and counterfactual Predictive Parity (cPP) as $Y^0 \perp A \mid \hat{Y} = \hat y$ where $\hat y = 0$ corresponds to negative predictive parity and $\hat y = 1$ corresponds to positive predictive parity.

\begin{theorem}[Predictive Parity]
Assume $\p(T=0 \mid y^0, a, \hat y) \neq 0$. If oPP holds, then cPP holds if and only if the following balance condition holds.
\begin{condition}[balPP]
\begin{equation} \label{eq:balPP}
     \begin{split}
         &\p(Y^1 =y \mid \hat y)\p(T=1 \mid Y^1 = y, \hat y) \\
         &- \p(Y^1 = y \mid a, \hat y)\p(T=1 \mid Y^1 =y, a, \hat y) \\&= \p(Y^0 = y \mid \hat y)\Big( \p(T=1 \mid Y^0 =y, \hat y) - \p(T=1 \mid Y^0 = y, a, \hat y) \Big)  
     \end{split}
     \end{equation}
     \end{condition}
     
BalPP is satisfied under the following independence conditions, which provide sufficient conditions for oPP to imply cPP.
\begin{condition}[indPP] 
\begin{equation}\label{eqn:indPP} 
\begin{split}
    T \perp A \mid  Y^0, \hat{Y} \\
     (Y^1, T) \perp A \mid \yh
\end{split}
\end{equation}  
\end{condition}
\end{theorem}

IndPP will not hold in many settings. Note that $(Y^1, T) \perp A \mid \yh \iff T \perp A \mid  Y^1, \yh $ and $Y^1 \perp A \mid \hat{Y}$.
\amdelete{
\begin{equation*}
    \begin{split}
        (Y^1, T) \perp A \mid \yh \iff &T \perp A \mid  Y^1, \yh \\
        &Y^1 \perp A \mid \hat{Y}
    \end{split} 
\end{equation*}} Conditions ${T \perp A \mid  Y^t, \yh}$ require 
$\yh$ to contain all the information that $A$ tells us about treatment assignment that is not contained in $Y^t$. Since $\yh$ is typically trained to predict $Y$ and not $T$, it is quite unlikely that these conditions will hold in settings where there is bias in treatment assignment even when controlling for true risk. Condition  ${Y^1 \perp A \mid \hat{Y}}$ allows differences in the risk distribution under treatment if we can fully explain these differences with $\yh$. In the best case $\yh \approx Y$, but it is unlikely that the observed outcome, which is not causally well-defined, would explain differences in the risk distribution under treatment. \amcomment{does this make sense? in my head it does because  you could imagine that conditioning on $Y^0$ could explain discrimination and conditioning on $Y^1$ certainly would, but conditioning on the observed outcome is a weird combination that no longer seems meaningful}
As above, even if indPP does not hold, balPP may hold 
but it is difficult to reason why this should hold in any setting.
Like Theorem 1, Theorem 2 also assumes a mild positivity-like assumption that is reasonable in risk assessment settings.


\subsubsection{Equalized odds} In settings where TPR and FPR are more important than predictive value, we may desire parity in TPR and FPR, a fairness notion known as Equalized Odds \cite{hardt2016equality}. Let observational Equalized Odds (oEO) require that $\hat{Y} \perp A \mid Y$ and counterfactual Equalized Odds (cEO) require that $\hat{Y} \perp A \mid Y^0$.

\begin{theorem}[Equalized Odds] 
Assume $\p(Y= y \mid a) \neq 0$ and $\p(T=0 \mid y^0, a, \hat y) \neq 0$. If oEO holds, then cEO holds if and only if the following balance condition holds.
.
 \begin{condition}[balEO]
  \begin{equation} \label{eq:balEO}
    \begin{split}
        &\p(\yh =1 \mid Y^1 = y)\frac{\p(T=1 \mid \yh = 1, Y^1 = y) \p(Y^1 = y)}{\p(Y = y)} \\
        &- \p(\yh =1 \mid Y^1 = y, a)\frac{\p(T=1 \mid \yh =1, Y^1 = y, a) \p(Y^1 = y \mid a)}{\p(Y = y \mid a)} \\
        &= \p(\yh=1 \mid Y^0 = y)\Bigg(
        \frac{\p(T=0 \mid \yh =1, Y^0 = y, a) \p(Y^0 = y \mid a)}{\p(Y = y \mid a)}\\
        &- \frac{\p(T=0 \mid \yh = 1, Y^0 = y) \p(Y^0 = y)}{\p(Y = y)} \Bigg)
    \end{split}
\end{equation}
 \end{condition}
 
 The balance condition is satisfied under the following independence conditions, which comprise sufficient conditions for oEO to imply cEO. 
  \begin{condition}[indEO]
  \begin{equation} \label{eqn:indEO}
      \begin{split}
          Y \perp A \\
          Y^0 \perp A \\
          T \perp A \mid \yh, Y^0 \\
         (Y^1, \yh, T) \perp A
      \end{split}
  \end{equation}
 \end{condition}
\end{theorem}

The first two conditions of indEO require oBP and cBP, so indEO requires balBP to hold.
In settings where there is discrimination in treatment assignment even when controlling for true risk, indEO is unlikely to hold. Even if there is no such discrimination, indEO will not hold if there are differences in the risk distributions under treatment since the last condition of~\ref{eqn:indEO}  requires $Y^1 \perp A$. indEO requires further conditions such as parity in the TPR/FPR against the outcome under treatment. If these conditions are not met, oEO could imply cEO if balEO holds, but it is difficult to reason about why this would hold for a setting when the independencies do not. Theorem 3 assumes two mild assumptions: the positivity-like assumption of Theorem 2 and $P(Y^0 = y \mid a) \neq 0$.
 
 Our theoretical analysis suggests that in many settings equalizing the observational fairness metric will not equalize the counterfactual fairness metric. We conclude by noting that the theorems hold when conditioning on any feature(s) $\subseteq X$, and in this context, these theorems are relevant to individual notions of fairness. 

\subsection{Experiments on synthetic data} \label{sec:fair_experiments}
We empirically demonstrate that equalizing the observational metric via fairness-corrective methods can increase disparity in the counterfactual metric 
on the synthetic data described in \textsection~\ref{sec:synth_risk}.\footnote{We do not perform the experiments on the child welfare data since it is balanced in terms of base rates and FPR/TPR with respect to race.}

\subsubsection{Reweighing} \label{sec:reweigh}
\begin{figure}
    \centering
    \includegraphics[scale=0.45, trim={0cm 0.65cm 0cm 0.25cm},clip]{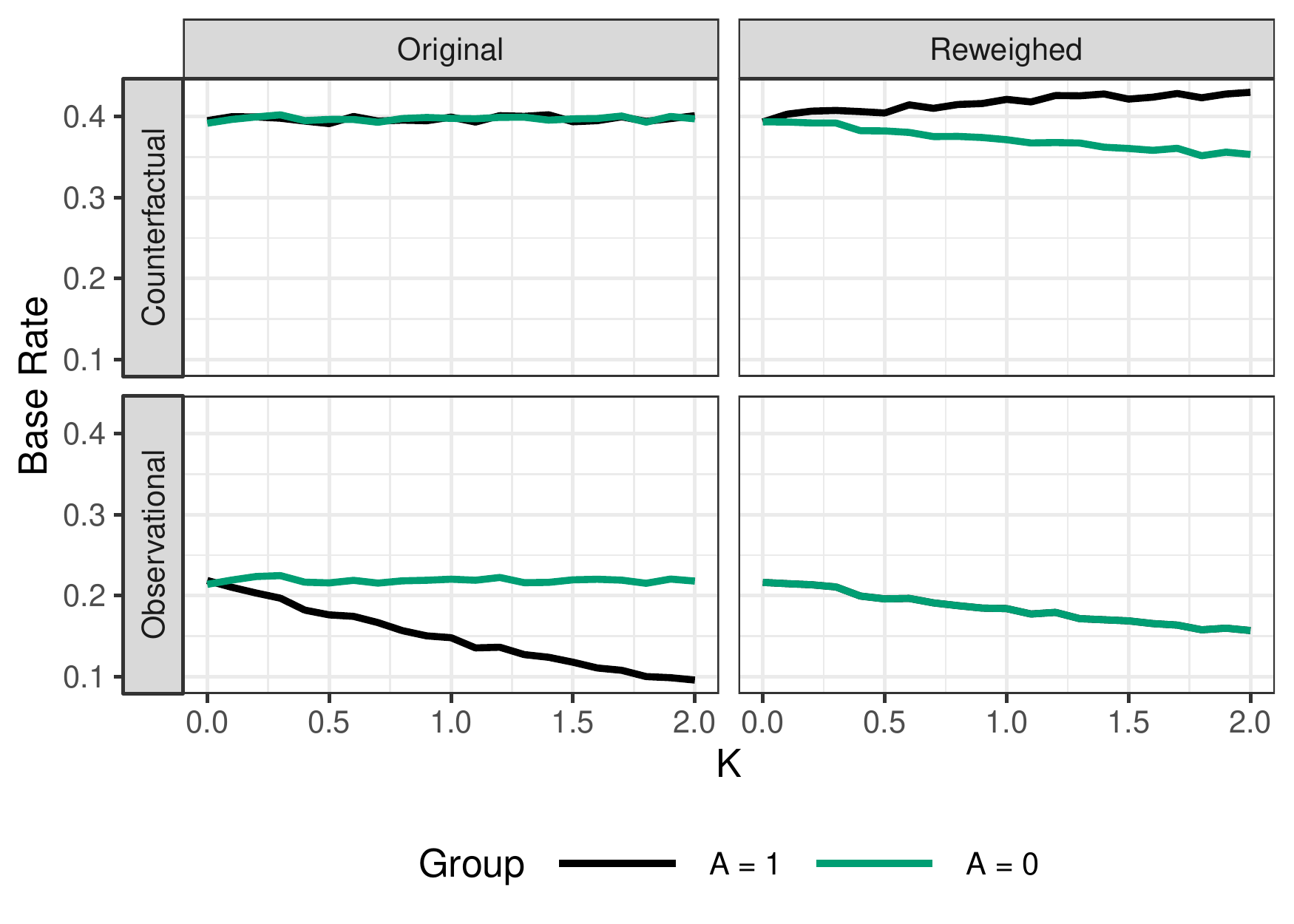}
    \caption{Counterfactual and observational base rates before and after applying a fairness-corrective method that reweighs training data (\textsection~\ref{sec:reweigh}). X-axis controls the bias of treatment assignment toward group $A=1$. Before reweighing (``Original"), counterfactual base rates are equal (cBP holds), but observational base rates are different (oBP doesn't hold) for $k >0$ since group $A=1$ is more likely to get treated. Reweighing achieves oBP but cBP no longer holds. } 
    \label{fig:synth_reweigh}
\end{figure}

One approach to encourage demographic parity reweighs the training data to achieve base rate parity \cite{kamiran2012data}.
 Figure ~\ref{fig:synth_reweigh} shows that without any processing (``Original"), the counterfactual base rates are equal while the observational base rates show increasing disparity with $K$. Reweighing applied to the observational outcome achieves oBP but induces disparity in the counterfactual base rate. Theorem ~\ref{thm:BP} suggested this result: For $k >0$, $A \not \perp T \mid Y^0$; then it is unlikely that oBP implies cBP.

\subsubsection{Post-processing for equalized odds} \label{sec:post}
We evaluate a  method that modifies scores to achieve a generalized version of equalized odds \cite{pleiss2017fairness, hardt2016equality}.\footnote{We use the Pleiss implementation on \url{https://github.com/gpleiss/equalized_odds_and_calibration} that extends the method in \cite{hardt2016equality} to probabilistic classifiers.} This method targets parity in the generalized FNR/FPR, where GFPR is $\e[\hat s(X) \mid Y=0]$ and GFNR is $\e[1-\hat s(X) \mid Y=1]$. We refer to these observational rates as oGFPR/oGFNR and define their counterfactual counterpart: cGFPR $= \e[\hat s(X) \mid Y^0=0]$ and cGFNR $= \e[1-\hat s(X) \mid Y^0=1]$.  We use the scores of the counterfactual model as inputs. We compute the cGFNR and cGFPR using our DR method from \textsection~\ref{sec:count_eval}.\footnote{The estimator is nearly identical to the estimators for FPR/FNR if we use $\hat s(X)$ in place of the predicted label $\hat Y(X)$} 

Table~\ref{fig:hardt_costs} shows that post-processing to equalize oGFPR and oGFNR induces imbalance in cGFPR and cGFNR.\footnote{We use $c=0.1$ and $k=1.6$. We report results for other values in Appendix~\ref{sec:appendix_fair_exp}.}
In Figure ~\ref{fig:hardt_roc} we see that the original model achieved cEO, but post-processing induced disparity to the detriment of the group that was less likely to be treated. Since treatment is beneficial,
this ``fairness" adjustment actually compounded the discrimination in the treatment assignment.
\begin{table}
    \centering
\begin{tabular}{llrrrr}
  \hline
Group & Method & cGFNR & cGFPR & oGFNR & oGFPR \\ 
  \hline
A=1 & Original & 0.50 & 0.33 & 0.58 & 0.39 \\ 
  A=0 & Original & 0.50 & 0.33 & 0.56 & 0.39 \\ 
  A=1 & Post-Proc. & 0.58 & 0.30 & 0.63 & 0.35 \\ 
  A=0 & Post-Proc. & 0.64 & 0.34 & 0.63 & 0.35 \\ 
   \hline
\end{tabular}
    \caption{Counterfactual and observational generalized FNR/FPR before and after post-processing to equalize odds (\textsection~\ref{sec:post}) using threshold $=0.5$. Before post-processing (``Original"), the counterfactual generalized rates (cGFNR and cGFPR) are the same for both groups. Post-processing equalizes the observational rates (oGFNR and oGFPR) but induces noticeable disparity in both cGFNR and cGFPR.}
    \label{fig:hardt_costs}
\end{table}

\begin{figure}
    \centering
    \includegraphics[scale=0.6, trim={0cm 0.65cm 0cm 0.5cm},clip]{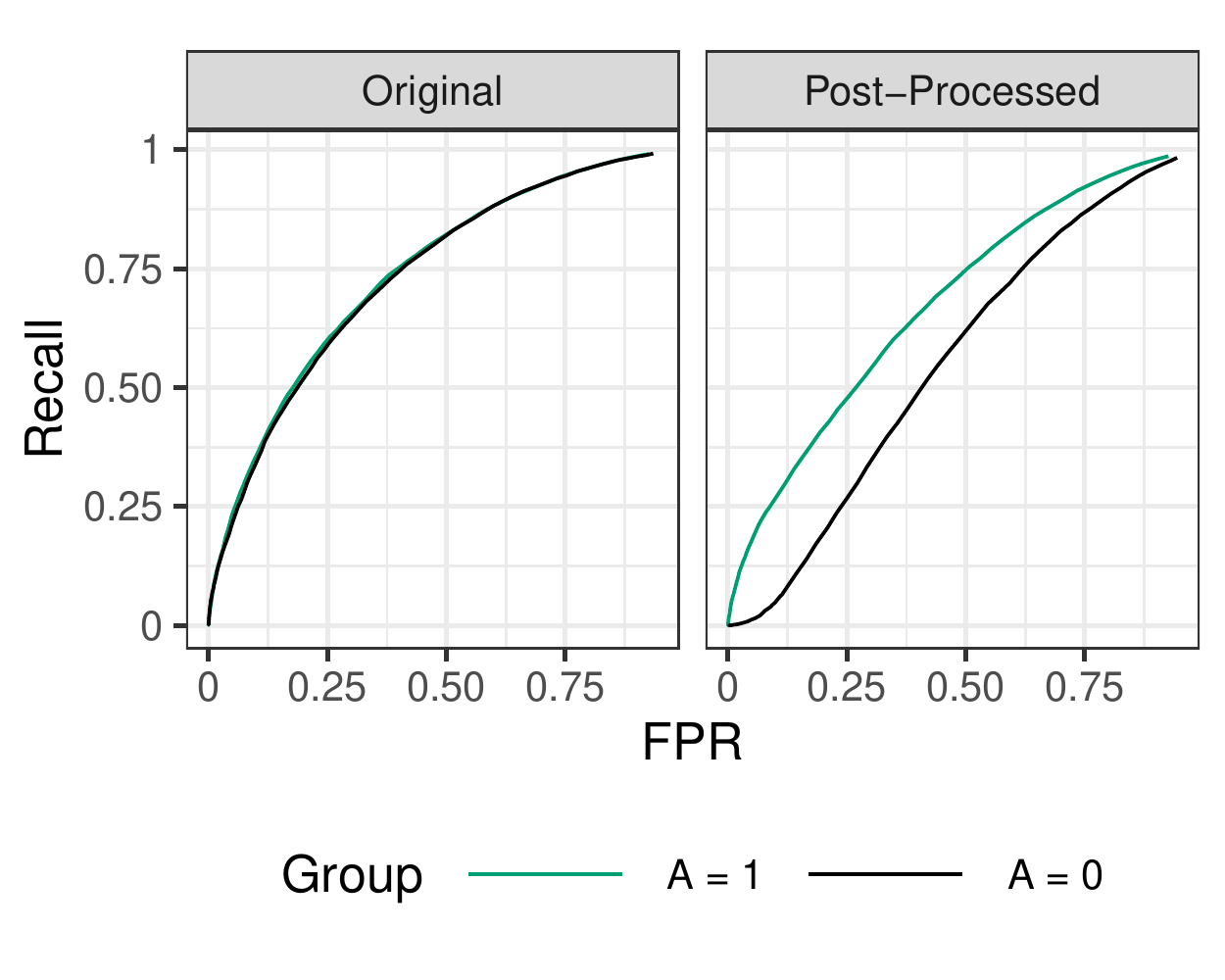}
    \caption{Counterfactual ROC curves before and after post-processing to equalize odds (\textsection~\ref{sec:post}). Before post-processing, ROC curves are identical for both groups, indicating that counterfactual equalized odds (cEO) holds. Post-processing induces imbalance, harming group $A=0$ and compounding initial unfairness in  treatment assignment.} 
    \label{fig:hardt_roc}
\end{figure}

\section{Conclusion}

This paper demonstrates that training and evaluating models using observed outcomes can lead to the misallocation of resources due to the misestimation of risk for those most receptive to treatment.  Furthermore, fairness-correcting methods that seek to achieve observational parity can lead to disparities on the relevant counterfactual metrics, and may further compound inequities in intial treatment assignment.   
The counterfactual approaches to learning, evaluation and predictive fairness assessment introduced in this paper provide more accurate and relevant indications of model performance.  

\begin{acks}
We are grateful to the Block Center for Technology and Society for funding this research.
This project would not have been possible without the support of Allegheny County Department of Human Services, who shared their data and answered many questions during the research process.
Thanks to our reviewers for helpful comments about the project.
\end{acks}

\clearpage
\bibliographystyle{ACM-Reference-Format}
\bibliography{ref}

\clearpage
\appendix
\section{Identifications}
In this section we give the identifications referenced in \textsection~\ref{sec:count_learning} and \textsection~\ref{sec:count_eval}.
Identification is the process of writing a counterfactual quantity in terms of observable quantities, based on causal assumptions.
Our identifications rely on the causal assumptions of \textsection~\ref{sec:learn} and assume that the model is learned and evaluated on separate train/test partitions (so the predictions $\yh$ are just a function of features $X$).

\subsection{Identification of the Counterfactual Outcome}
In \textsection~\ref{sec:count_learning}, we identify the counterfactual target $\e[Y^0 \mid X] = \e[Y \mid X, T =0] $. The derivation is 
\begin{eqnarray*}
\e[Y^0 \mid X] &= \e[Y^0 \mid X, T =0] \\
&= \e[Y \mid X, T =0] 
\end{eqnarray*}
where the first line used the exchangeability assumption in \textsection~\ref{sec:count_learning} and the second line used the consistency assumption in \textsection~\ref{sec:count_learning}.

\subsection{Identification of Counterfactual TPR}
In \textsection~\ref{sec:count_eval}, we identify the counterfactual TPR (or recall) as \begin{equation*}
     {\mathbb{E}[\hat{Y} \mid Y^0 = 1]} = \frac{ \mathbb{E}\big[\hat{Y}\mathbb{E}[Y \mid X,  T = 0]\big]}{\mathbb{E} \Big[ \mathbb{E} [ Y \mid X, T = 0]\Big]} 
\end{equation*}

The derivation is as follows. By definition of conditional expectation we have
\begin{eqnarray*}
{\mathbb{E}[\hat{Y} \mid Y^0 = 1]} = \frac{\e[\hat Y \mathbb{I} \{ Y^0 = 1 \}]}{\p(Y^0 = 1)} 
\end{eqnarray*}

We separately identify the numerator and denominator. 
Since we are evaluating on a test partition, $\hat Y$ is a function of $X$. Then for the numerator we have 
\begin{eqnarray*}
\e[\hat Y \mathbb{I} \{ Y^0 = 1 \}] &= \e[ \hat Y  \e [\mathbb{I} \{ Y^0 = 1 \}\mid X]] \\
&= \e[ \hat Y  \p (Y^0 = 1 \mid X)] \\
&= \e[ \hat Y  \e [Y^0 \mid X]] \\
&= \e[ \hat Y  \e [Y^0 \mid X, T=0]] \\
&= \e[ \hat Y  \e [Y \mid X, T=0]] \\
\end{eqnarray*}
where the first line used iterated expectation, the second line used the definition of an indicator function, the third line use the fact that $Y^0 \in \{0,1\}$, the fourth line used exchangeability, and the fifth line used consistency.

To identify the denominator, we use iterated expectation and then apply exchangeability and consistency as we did for the counterfactual target:
\begin{eqnarray*}
\p(Y^0 = 1) &= \e[Y^0] \\
&= \e[ \e[Y^0 \mid X]] \\
&= \e[ \e[Y^0 \mid X, T =0]] \\
&= \e[ \e[Y \mid X, T =0]] \\
\end{eqnarray*}

\subsection{Identification of Counterfactual Precision}
In \textsection~\ref{sec:count_eval}, we identify the counterfactual precision as 

\begin{equation*}
    \mathbb{E}[Y^0 \mid \hat{Y} = 1] = \mathbb{E}[\mathbb{E}[Y \mid X, T = 0] \mid \hat{Y} =1 ]
\end{equation*}. 

The derivation is as follows. 
\begin{eqnarray*}
    \mathbb{E}[Y^0 \mid \hat{Y} =1] &= \e[\e[Y^0 \mid X, \hat{Y} = 1] \mid \yh = 1] \\
    &= \e[\e[Y^0 \mid X] \mid \yh = 1] \\
    &= \e[\e[Y^0 \mid X, T =0] \mid \yh = 1] \\
    &= \e[\e[Y \mid X, T =0] \mid \yh = 1] \\
\end{eqnarray*}.

where the first line uses iterated expected, the second line applies the fact that $\yh$ is just a function of $X$, the third line uses exchangeability (from \textsection~\ref{sec:count_learning}) and the last lines uses consistency (from \textsection~\ref{sec:count_learning}).

\subsection{Identification of Counterfactual Calibration}
The derivation is the same as for precision since $\hat s(X)$ is just a function of $X$.

\subsection{Identification of Counterfactual FPR}
In \textsection~\ref{sec:count_eval}, we identified counterfactual FPR as 

\begin{equation*}
    \mathbb{E}[\hat{Y} \mid Y^0 = 0] = \frac{ \mathbb{E}\Big[\hat{Y} \mathbb{E}[1-Y \mid X,  T = 0]\Big]}{\mathbb{E} \Big[ \mathbb{E} [ 1-Y \mid X, T = 0]\Big]}  
\end{equation*}

The below derivation is similar to that of TPR.
We can rewrite the target as
\begin{eqnarray*}
\mathbb{E}[\hat{Y} \mid Y^0 = 0] &= \frac{\e[\yh \mathbb{I}\{ Y^0 = 0\}]}{\p(Y^0 = 0)}
\end{eqnarray*}

We separately identify the numerator and denominator. For the numerator we have 
\begin{eqnarray*}
\e[\yh \mathbb{I}\{ Y^0 = 0\}] &= \e[\yh \e [\mathbb{I}\{ Y^0 = 0\} \mid X]] \\
&= \e[\yh \p( Y^0 = 0 \mid X) ] \\
&= \e[\yh 1-\e[Y^0 \mid X] ] \\ 
&= \e[\yh \e[1-Y^0 \mid X, T = 0] ] \\
&= \e[\yh \e[1-Y \mid X, T = 0] ] \\
\end{eqnarray*}
where the first line used iterated expectation, the second line used the definition of indicator function, the third line used the fact that $Y^0$ is a binary random variable, the fourth line used exchangeability, and the last line used consistency.

For the denominator, we have 
\begin{eqnarray*}
\p(Y^0 = 0) &= 1 - \p(Y^0 =1) \\
&= 1- \e[ \e[Y \mid X, T = 0]] \\
&=\e[ \e[1-Y \mid X, T=0]]
\end{eqnarray*}
where the second line used the derivation for the denominator in TPR.

\section{Proofs}\label{appendix_proofs}
In this section we give the proofs for theorems in \textsection~\ref{sec:fair_theorems}. These proofs assume consistency (defined in \textsection~\ref{sec:count_learning}).

\begin{proof} [Proof that balBP is Necessary and Sufficient]
 By consistency $Y = T Y^1 + (1-T) Y^0$. Then we have $\mathbb{P}(Y = y)=$ 
 \begin{equation*}\label{eqn:nBPo} \mathbb{P}(Y^1= y)\mathbb{P}(T= 1 \mid Y^1=y) +\mathbb{P}(Y^0 =y)\mathbb{P}(T= 0 \mid Y^0 =y) 
 \end{equation*}
 Likewise for $\mathbb{P}(Y = y \mid a) =$ 
 \begin{equation*} \label{eqn:nBPa}
    \begin{split}
    &\mathbb{P}(Y^1 = y\mid a)\mathbb{P}(T= 1 \mid Y^1 =y, a) +\mathbb{P}(Y^0 =y \mid a)\mathbb{P}(T= 0 \mid Y^0 = y, a) 
    \end{split}
 \end{equation*}
By oBP, $\mathbb{P}(Y = y)= \mathbb{P}(Y = y \mid a)$. By the above expansions,
\begin{equation}\label{eqn:oBP_expans}
\begin{split}
&\mathbb{P}(Y^1 =y)\mathbb{P}(T= 1 \mid Y^1=y) -\mathbb{P}(Y^1=y \mid a)\mathbb{P}(T= 1 \mid Y^1 =y, a) \\
    &= \mathbb{P}(Y^0=y \mid a)\mathbb{P}(T= 0 \mid Y^0=y, a) -\mathbb{P}(Y^0 =y)\mathbb{P}(T= 0 \mid Y^0=y) 
    \end{split}
 \end{equation}
\emph{Necessary:} For oBP to imply cBP, both conditions must hold. By cBP,  $\mathbb{P}(y^0) = \mathbb{P}(y^0 \mid a)$.  Equation~\ref{eqn:oBP_expans} then becomes 
  \begin{equation} \label{eqn:balBP_proof}
     \begin{split}
         &\p(Y^1 = y)\p(T=1 \mid Y^1 = y) - \p(Y^1 =y \mid a)\p(T=1 \mid Y^1 =y, a) \\&= \p(Y^0 =y)\Big( \p(T=0 \mid Y^0 = y, a) - \p(T=0 \mid Y^0 = y) \Big)
     \end{split}
 \end{equation}
 which is the balBP condition since we can rewrite the right-hand side as $\p(Y^0 = y)\Big( \p(T=1 \mid Y^0 = y) - \p(T=1 \mid Y^0 = y, a) \Big)$.
 
 \emph{Sufficient:} In addition to oBP, we assume balBP holds. The left-hand sides of balBP (Equation ~\ref{eqn:balBP_proof}) and oBP (Equation ~\ref{eqn:oBP_expans}) are the same. Then applying the transitive property, 
 \begin{equation*} \label{eq:BR}
     \begin{split}
     &\p(Y^0 =y)\Big( \p(T=0 \mid Y^0 = y, a) - \p(T=0 \mid Y^0 = y) \Big)
          \\&=  \mathbb{P}(Y^0 = y \mid a)\mathbb{P}(T= 0 \mid Y^0 =y, a) -\mathbb{P}(Y^0 = y)\mathbb{P}(T= 0 \mid Y^0 =y)  
     \end{split}
 \end{equation*}
 
 Assuming $\p(T=0 \mid y^0, a) \neq 0$ (which is a mild positivity-like assumption), we conclude that $\p(y^0 \mid a) = \p(y^0) \implies Y^0 \perp A$. 
\end{proof}

\paragraph{indBP Sufficiency}
The following conditions are sufficient for oBP to imply cBP: $T \perp A \mid  Y^0$ and  $(Y^1, T) \perp A$
\begin{proof}[Proof of Base Rate Parity Sufficiency]
 If indBP holds, then balBP holds: Since $T \perp A \mid  Y^0$, the LHS of Equation~\ref{eqn:balBP_proof} is zero. By contraction  $(Y^1, T) \perp A$ is equivalent to  $Y^1 \perp A$ and  $T \perp A \mid Y^1$; then the RHS of Equation~\ref{eqn:balBP_proof} is also zero, so balBP holds. Since balBP is sufficient, indBP is sufficient for oBP to imply cBP.
 \end{proof}

\paragraph{Predictive Parity}
The proofs use the same techniques as for base rate parity.

\begin{proof}[Proof that BalEO is Necessary and Sufficient]

We first expand $\p(\yh =1 \mid Y = y)$
\begin{eqnarray}
&= \frac{\p(\yh =1, Y =y)}{\p(Y=y)} \\
&= \frac{\p(\yh =1, Y^1 =y, T = 1) + \p(\yh =1, Y^0 = y, T = 0)}{\p(Y = y)} \\
\end{eqnarray}

which we can further expand to get  $\p(\yh =1 \mid Y = y)$
\begin{equation} \label{eqn:eo_expansion}
    \begin{split}
        &= \frac{\p(T=1 \mid \yh =1, Y^1 =y) \p(\yh=1 \mid Y^1 = y) \p(Y^1=y)}{\p(Y=y)} \\ &+ \frac{\p(T=0 \mid \yh =1, Y^0 = y) \p(\yh=1 \mid Y^0 = y) \p(Y^0 = y)}{\p(Y = y)}  \\
    \end{split}
\end{equation}
 
Since oEO holds by assumption, then $\p(\yh =1 \mid Y = y) = \p(\yh =1 \mid Y = y, A = a)$. Using the expansion in Equation~\ref{eqn:eo_expansion}, we have 
\begin{equation}
    \begin{split}
        & \frac{\p(T=1 \mid \yh =1, Y^1 = y) \p(\yh=1 \mid Y^1 = y) \p(Y^1=y)}{\p(Y = y)} \\ &+ \frac{\p(T=0 \mid \yh =1, Y^0 = y) \p(\yh=1 \mid Y^0 = y) \p(Y^0 = y)}{\p(Y = y)}  \\
        &= \frac{\p(T=1 \mid \yh =1, Y^1 = y, a) \p(\yh=1 \mid Y^1 = y, a) \p(Y^1=y \mid a)}{\p(Y=y \mid a)} \\ &+ \frac{\p(T=0 \mid \yh =1, Y^0 = y, a) \p(\yh=1 \mid Y^0 = y, a) \p(Y^0= y \mid a)}{\p(Y = y \mid a)}  \\
    \end{split}
\end{equation}

Rearranging gives
\begin{equation} \label{eqn:oEO_implies}
    \begin{split}
        & \p(\yh=1 \mid Y^1 = y) \frac{\p(T=1 \mid \yh =1, Y^1 = y)  \p(Y^1=y)}{\p(Y=y)} \\&- \p(\yh=1 \mid Y^1 = y, a)\frac{\p(T=1 \mid \yh =1, Y^1 = y, a)  \p(Y^1=y \mid a)}{\p(Y = y \mid a)} \\&=- \p(\yh=1 \mid Y^0 = y)\frac{\p(T=0 \mid \yh =1, Y^0 = y)  \p(Y^0 = y)}{\p(Y = y)}  \\&+ \p(\yh=1 \mid Y^0 = y, a) \frac{\p(T=0 \mid \yh =1, Y^0 = y, a)  \p(Y^0 = y \mid a)}{\p(Y = y \mid a)} 
    \end{split}
\end{equation}

\paragraph{Necessary} For oEO to imply cEO, both conditions must hold. By cEO, $\p(\yh =1 \mid Y^0 = y) = \p(\yh =1 \mid Y^0 = y, A = a)$  which would imply that 
\begin{equation} \label{eqn:balEO_statement}
    \begin{split}
        &\p(\yh =1 \mid Y^1 = y)\frac{\p(T=1 \mid \yh = 1, Y^1 = y) \p(Y^1 = y)}{\p(Y = y)} \\
        &- \p(\yh =1 \mid Y^1 = y, a)\frac{\p(T=1 \mid \yh =1, Y^1 = y, a) \p(Y^1 = y \mid a)}{\p(Y = y \mid a)} \\
        &= \p(\yh=1 \mid Y^0 = y)\Bigg(
        \frac{\p(T=0 \mid \yh =1, Y^0 = y, a) \p(Y^0 = y \mid a)}{\p(Y = y \mid a)}\\
        &- \frac{\p(T=0 \mid \yh = 1, Y^0 = y) \p(Y^0 = y)}{\p(Y = y)} \Bigg)
    \end{split}
\end{equation}

\paragraph{Sufficient} In addition to oEO, we assume balEO holds. From oEO we have equation~\ref{eqn:oEO_implies} and balEO is equation~\ref{eqn:balEO_statement}. The left-hand sides of equations~\ref{eqn:oEO_implies} and~\ref{eqn:balEO_statement} are the same so by the transitive property, 
\begin{equation}
    \begin{split}
        &- \p(\yh=1 \mid Y^0 = y)\frac{\p(T=0 \mid \yh =1, Y^0 = y)  \p(Y^0 = y)}{\p(Y = y)}  \\&+ \p(\yh=1 \mid Y^0 = y, a) \frac{\p(T=0 \mid \yh =1, Y^0 = y, a)  \p(Y^0 = y \mid a)}{\p(Y = y \mid a)} \\
        &= \p(\yh=1 \mid Y^0 = y)\Bigg(
        \frac{\p(T=0 \mid \yh =1, Y^0 = y, A = a) \p(Y^0 = y \mid a)}{\p(Y = y \mid a)}\\
        &- \frac{\p(T=0 \mid \yh = 1, Y^0 = y) \p(Y^0 = y)}{\p(Y = y)} \Bigg)
    \end{split}
\end{equation}

Simplifying gives 
\begin{equation}
    \begin{split}
        &\p(\yh=1 \mid Y^0 = y, a) \frac{\p(T=0 \mid \yh =1, Y^0 = y, a)  \p(Y^0 = y \mid a)}{\p(Y = y \mid a)} \\
        &= \p(\yh=1 \mid Y^0 = y)\Bigg(
        \frac{\p(T=0 \mid \yh =1, Y^0 = y, a) \p(Y^0 = y \mid a)}{\p(Y = y \mid a)} \Bigg)
    \end{split}
\end{equation}

Assuming $\p(T=0 \mid \yh =1, y^0, a) \neq 0$ and $ \p(Y = y \mid a) \neq 0$, then we conclude that $\p(\yh=1 \mid Y^0 = y, a)=\p(\yh=1 \mid Y^0 = y) \implies cEO$ 

\end{proof}

\paragraph{indEO Sufficiency}
The following conditions are sufficient for oEO to imply cEO: \begin{equation}
      \begin{split}
          Y \perp A \\
          Y^0 \perp A \\
          T \perp A \mid \yh, Y^0 \\
         (Y^1, \yh, T) \perp A
      \end{split}
  \end{equation}
\begin{proof}[Proof of indEO Sufficiency]
By contraction, the indEO conditions are equivalently written as $\yh \perp A \mid Y^1$;
$Y \perp A$; $Y^1 \perp A$; $Y^0 \perp A$; $T \perp A \mid \yh, Y^0$; $T \perp A \mid \yh, Y^1$. 
Under these assumptions, both sides of Equation~\ref{eqn:balEO_statement} are 0, so the balEO condition holds under these independencies. Since balEO is sufficient, then indEO is sufficient for oEO to imply cEO.
\end{proof}

\clearpage

\clearpage
\section{Child Welfare Evaluation via Task adaptation on Services}\label{sec:appendix_cw}
In \textsection~\ref{sec:cw_task}, we assessed how well our models of risk performed on the related risk task of predicting out-of-home placement. 
Another related risk task considers the decision to accept a case for services.
A case can only be accepted for services if it is under investigation, and we assume that the child welfare process assigns services based on the assessment of risk of child harm. 
Table \ref{fig:au-services} shows the AuROC and AuPR for the services task. As for the placement task, the observational model performs worse than random, and the counterfactual model performs better than random.
\begin{table}[ht]
\centering
\begin{tabular}{rllr}
  \hline
 & Observational & Counterfactual & Random \\ 
  \hline
AUROC & 0.40 (0.39,0.41) & 0.63 (0.61,0.67) & 0.50 \\ 
  AUPR & 0.33 (0.32,0.35) & 0.49 (0.47,0.50) & 0.41 \\ 
   \hline
\end{tabular}
\caption{Area under ROC and PR curves using our re-referral models to predict a related risk task, accepting the case for services (95\% confidence intervals given in parentheses). The observational model performs worse than a random classifier. The counterfactual model performs better; it learns a model of risk that transfers to related risk tasks whereas the observational model does not.} 
\label{fig:au-services}
\end{table}
\clearpage

\section{Further Synthetic Evaluation}\label{sec:appendix_synthetic}
We supplement the empirical analysis in \textsection~\ref{sec:synth_risk} by  presenting the PR, ROC, and calibration curves for several values of $c$, the parameter describing treatment effect, and $k$, the parameter describing treatment assignment bias. Each column denotes an evaluation method as described in Section ~\ref{sec:eval}.


\begin{figure*}
\centering
\begin{subfigure}{\textwidth}
  \centering
  \includegraphics[scale=0.5, trim={0cm 0.5cm 0cm 0cm},clip]{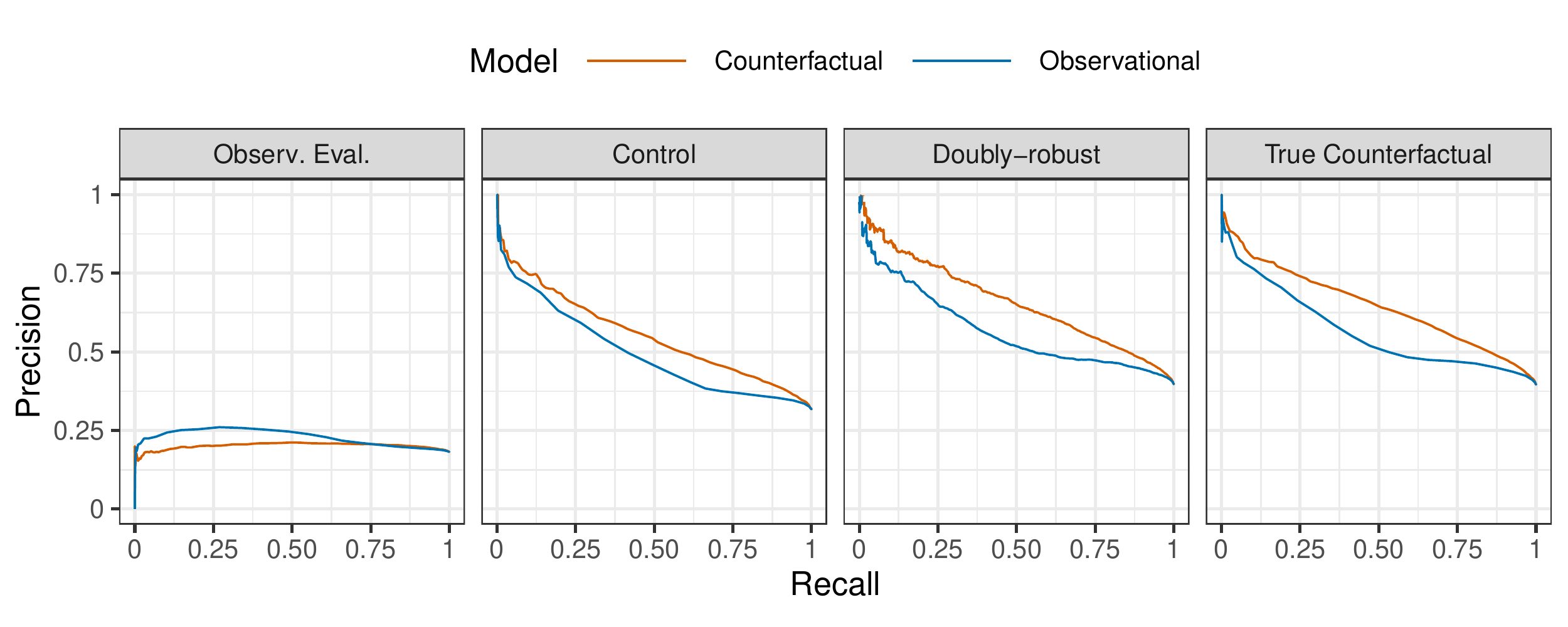}
    \caption{PR curves}
    \label{fig:synth_pr_c10_k100}
\end{subfigure}%
\\
\begin{subfigure}{\textwidth}
  \centering
  \includegraphics[scale=0.5, trim={0cm 0.6cm 0cm 0cm},clip]{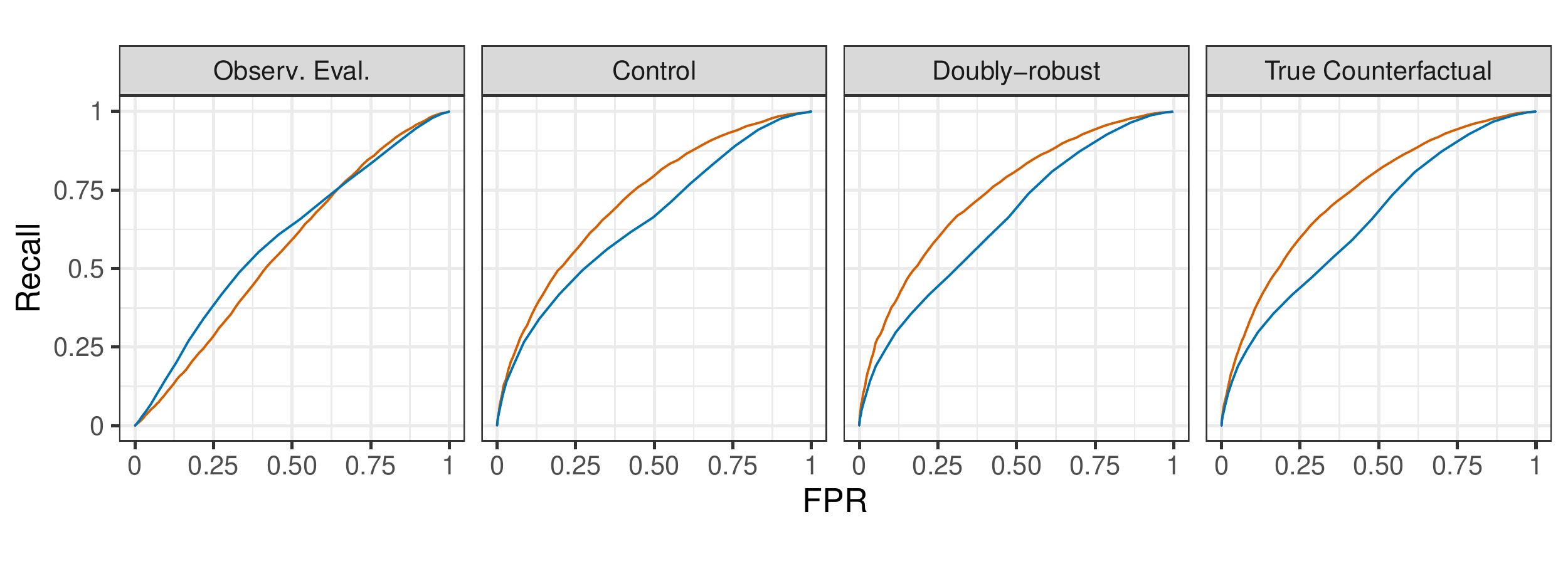}
  \caption{ROC curves}
  \label{fig:synth_roc_c10_k100}
\end{subfigure}
\\
\begin{subfigure}{\textwidth}
  \centering
  \includegraphics[scale=0.5, trim={0cm 0.6cm 0cm 0cm},clip]{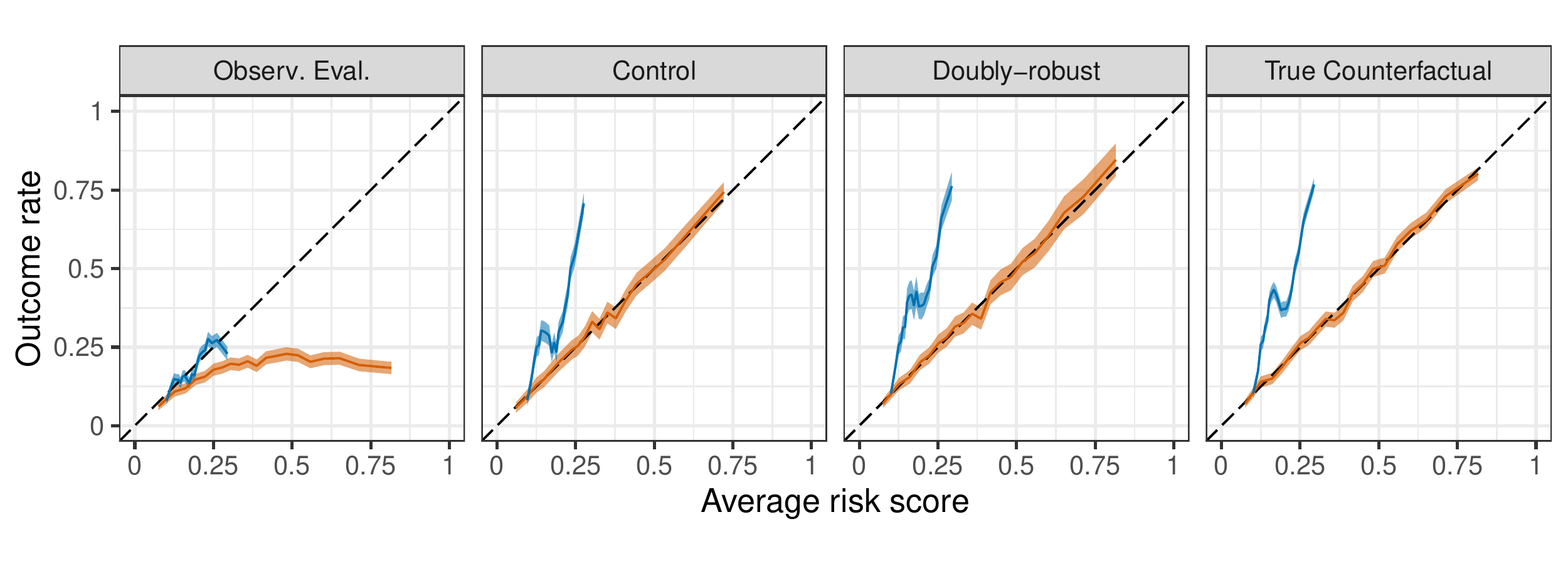}
  \caption{Calibration curves. 95\% pointwise confidence bounds shown.}
  \label{fig:synth_calib_c10_k100}
\end{subfigure}
\caption{Synthetic data results. with parameters with $c =0.1$, $k =1$. 
Each column pertains to a different evaluation method (described in \textsection~\ref{sec:eval}). Colors denote the learning method (described in \textsection~\ref{sec:learn}). Comparing to Figure~\ref{fig:synth} where $k=1.6$, a smaller value of $k$ here reduces the propensity to treat group $A=1$. This reduces treatment imbalance and treatment rates, and therefore the observational model performs better on this data than the data generated with $k=1.6$ but still underperforms compared to the counterfactual model. As in Figure~\ref{fig:synth}, the DR evaluation most accurately represents the true counterfactual evaluation. The observational evaluation incorrectly suggests the observational model outperforms the counterfactual model. The control evaluation gives inaccurate curves. (See \textsection~\ref{sec:synth_risk} for details on the data generation)
}
\end{figure*}

\begin{figure*}
\centering
\begin{subfigure}{\textwidth}
  \centering
  \includegraphics[scale=0.5, trim={0cm 0.5cm 0cm 0cm},clip]{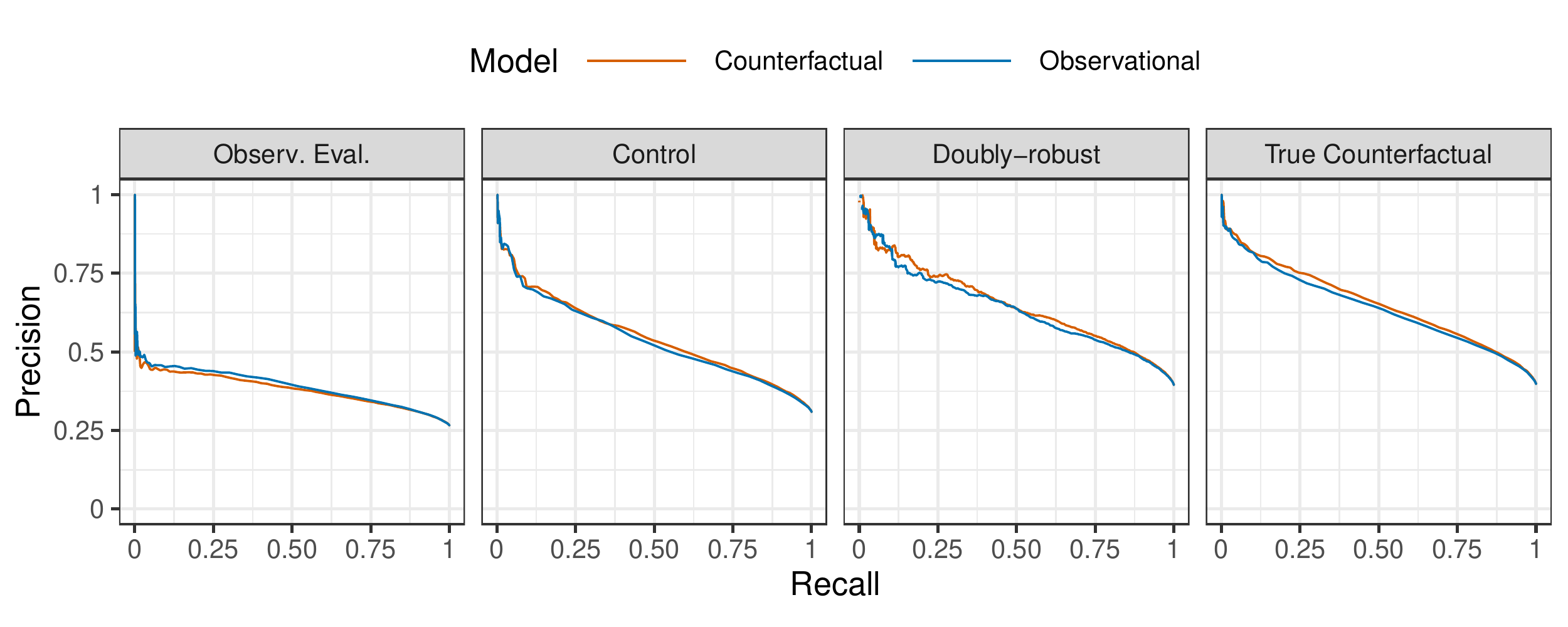}
    \caption{PR curves}
    \label{fig:synth_pr_c50_k160}
\end{subfigure}%
\\
\begin{subfigure}{\textwidth}
  \centering
  \includegraphics[scale=0.5, trim={0cm 0.6cm 0cm 0cm},clip]{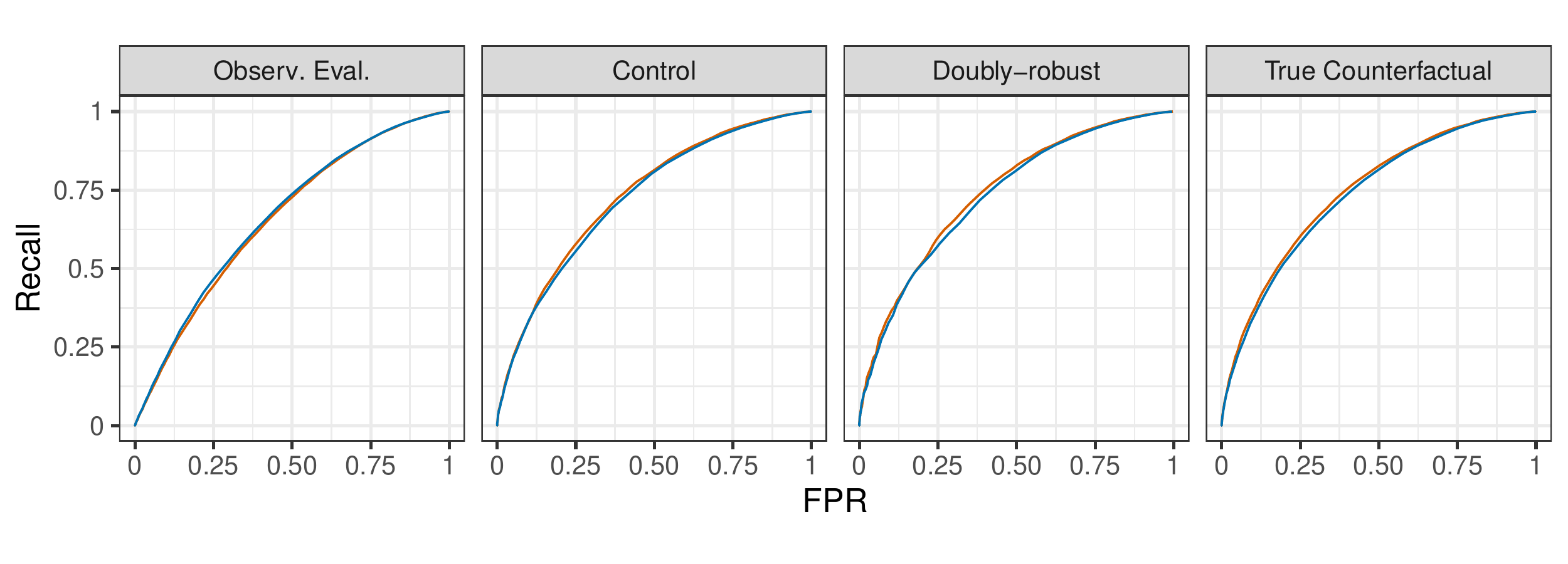}
  \caption{ROC curves}
  \label{fig:synth_roc_c50_k160}
\end{subfigure}
\\
\begin{subfigure}{\textwidth}
  \centering
  \includegraphics[scale=0.5, trim={0cm 0.6cm 0cm 0cm},clip]{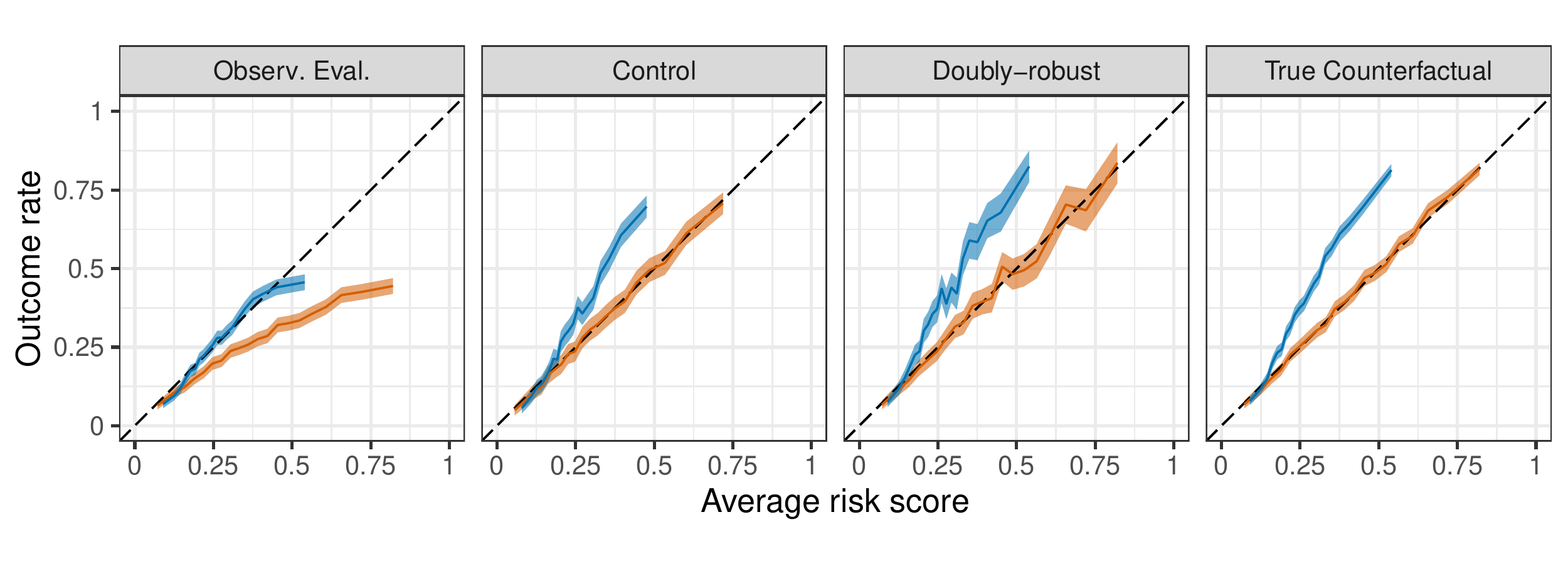}
  \caption{Calibration curves. 95\% pointwise confidence bounds shown.}
  \label{fig:synth_calib_c50_k160}
\end{subfigure}
\caption{Synthetic data results. with parameters with $c =0.5$, $k =1.6$. Each column pertains to a different evaluation method (described in \textsection~\ref{sec:eval}). Colors denote the learning method (described in \textsection~\ref{sec:learn}). Comparing to Figure~\ref{fig:synth} where $c= 0.1$, a higher value of $c$ here  decreases the treatment effect and correspondingly the amount the counterfatual model outperforms the observational model. The PR and ROC curves for the counterfactual and observational models are much closer than in Figure~\ref{fig:synth}. The counterfactual model still outperforms in terms of calibration. Notably, the control and observational evaluations give incorrect calibration and PR curves. The DR evaluation most closely resembles the true counterfactual evaluation.}
\end{figure*}

\begin{figure*}
\centering
\begin{subfigure}{\textwidth}
  \centering
  \includegraphics[scale=0.5, trim={0cm 0.5cm 0cm 0cm},clip]{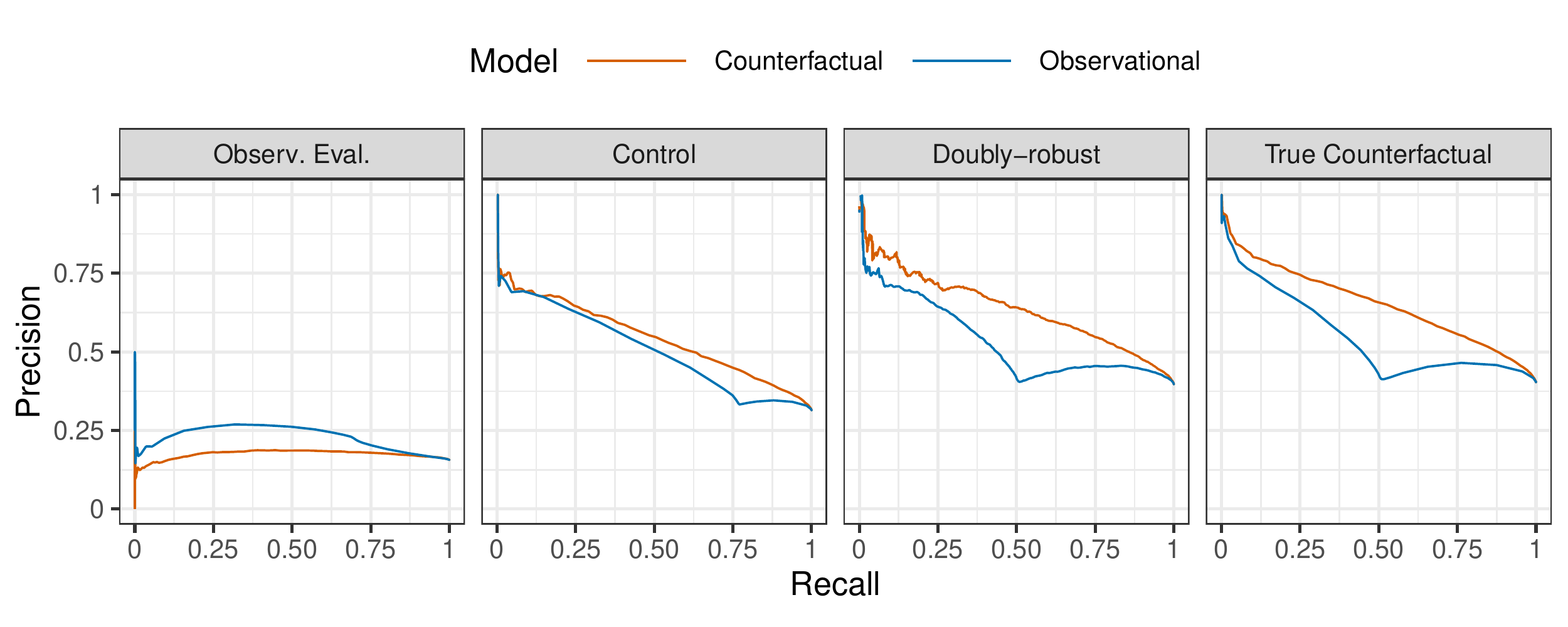}
    \caption{PR curves}
    \label{fig:synth_pr_c10_k200}
\end{subfigure}%
\\
\begin{subfigure}{\textwidth}
  \centering
  \includegraphics[scale=0.5, trim={0cm 0.6cm 0cm 0cm},clip]{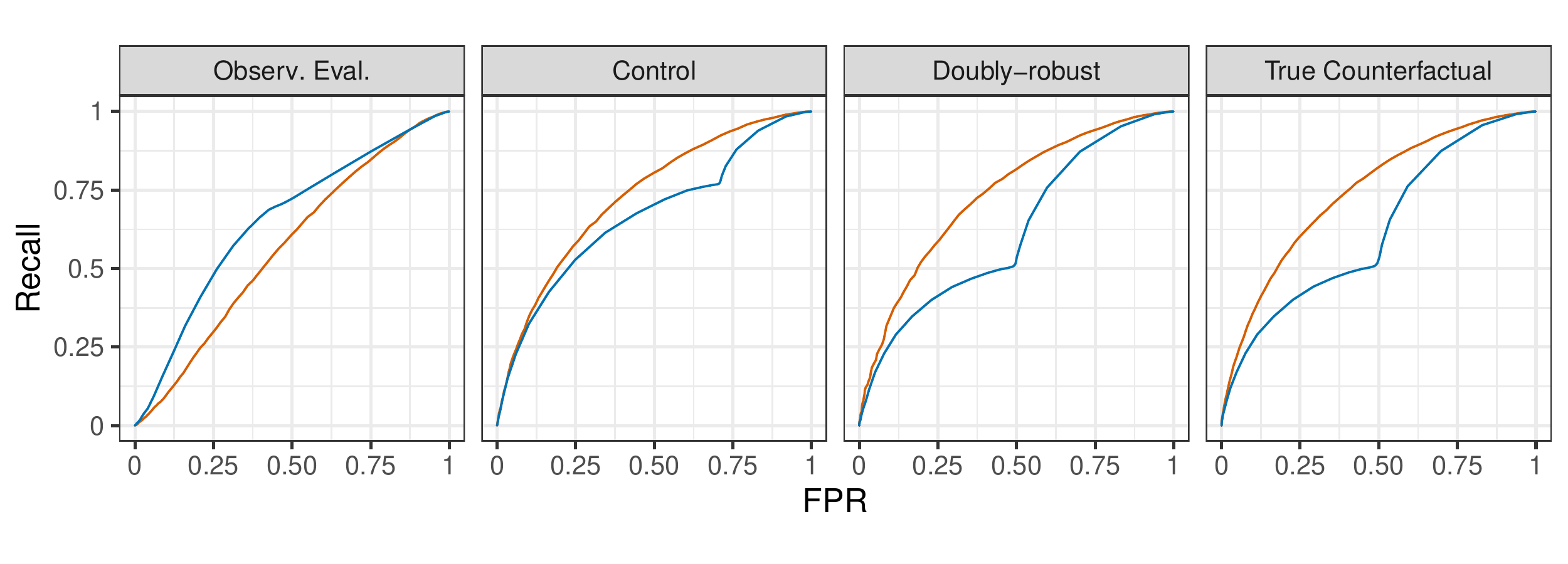}
  \caption{ROC curves}
  \label{fig:synth_roc_c10_k200}
\end{subfigure}
\\
\begin{subfigure}{\textwidth}
  \centering
  \includegraphics[scale=0.5, trim={0cm 0.6cm 0cm 0cm},clip]{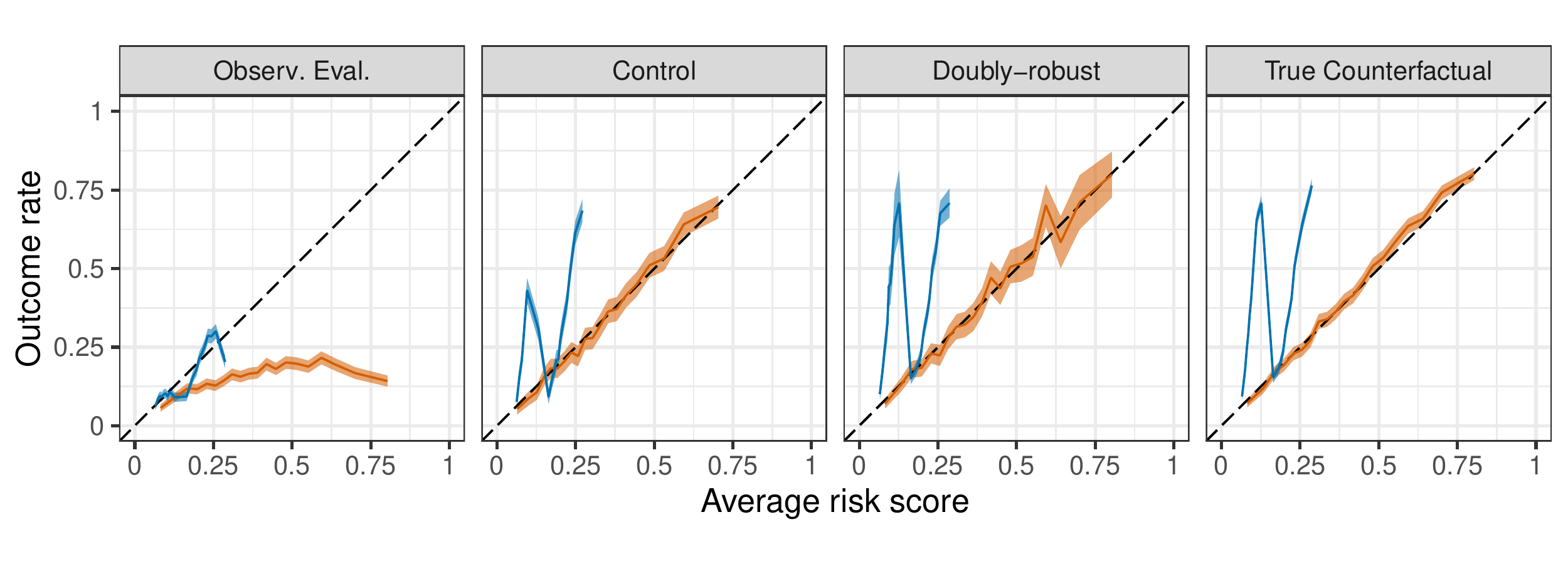}
  \caption{Calibration curves. 95\% pointwise confidence bounds shown.}
  \label{fig:synth_calib_c10_k200}
\end{subfigure}
\caption{Synthetic data results. with parameters with $c =.1$, $k =2$. Each column pertains to a different evaluation method (described in \textsection~\ref{sec:eval}). Colors denote the learning method (described in \textsection~\ref{sec:learn}). Comparing to Figure~\ref{fig:synth} where $k=1.6$, a larger value of $k$ here increases the propensity to treat group $A=1$. This increases treatment imbalance and treatment rates. The DR evaluation most accurately represents the true counterfactual evaluation. }
\end{figure*}
\clearpage

\subsection{Controlling for Treatment in the Regression} \label{sec:appendix_synth_treat}
\begin{figure*}
\centering
\begin{subfigure}{\textwidth}
  \centering
  \includegraphics[scale=0.5, trim={0cm 0.5cm 0cm 0cm},clip]{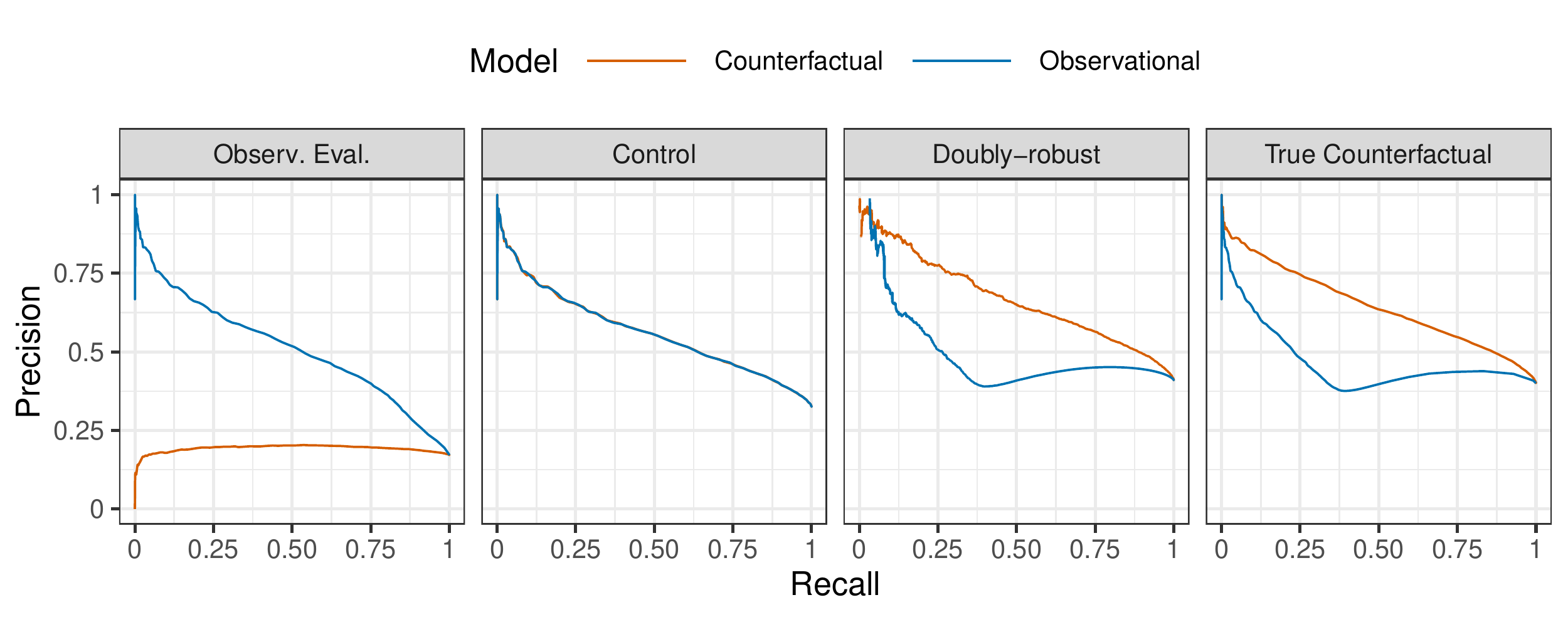}
    \caption{PR curves}
\end{subfigure}%
\\
\begin{subfigure}{\textwidth}
  \centering
  \includegraphics[scale=0.5, trim={0cm 0.6cm 0cm 0cm},clip]{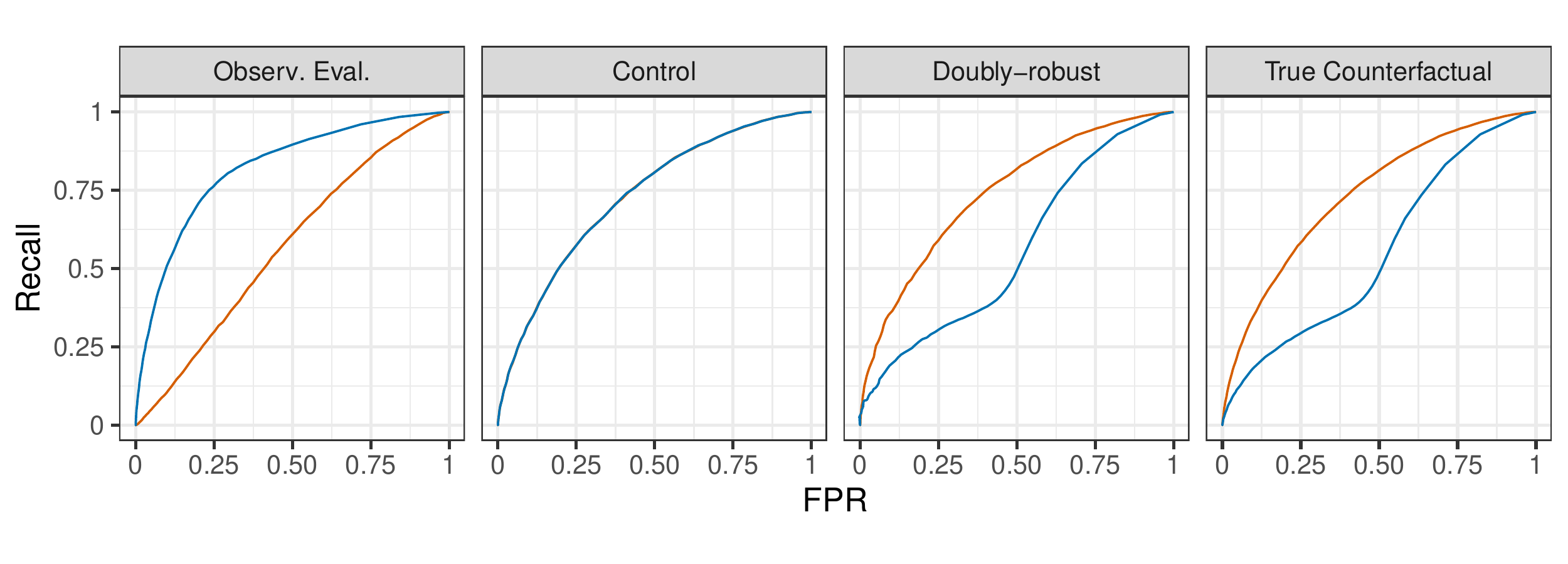}
  \caption{ROC curves}
\end{subfigure}
\\
\begin{subfigure}{\textwidth}
  \centering
  \includegraphics[scale=0.5, trim={0cm 0.6cm 0cm 0cm},clip]{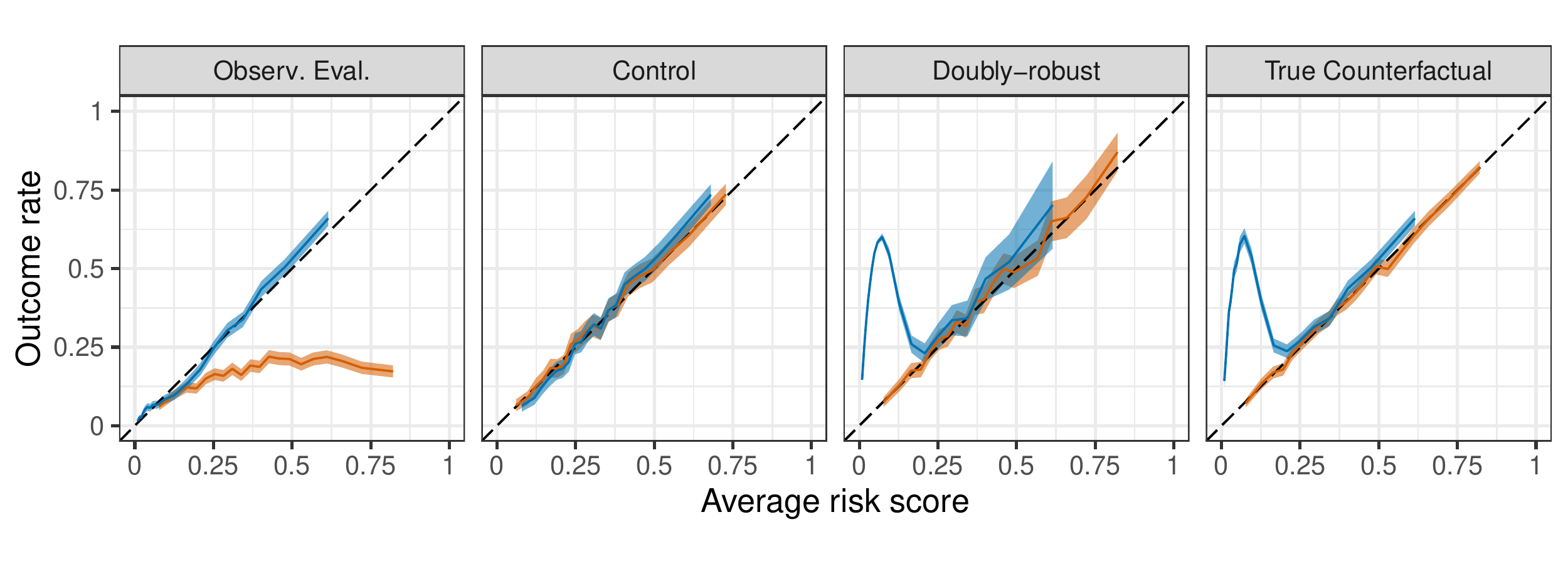}
  \caption{Calibration curves. 95\% pointwise confidence bounds shown.}
  \label{fig:synth_calib_treat_c10_k160}
\end{subfigure}

\caption{Synthetic data results when including treatment decision as a feature with parameters with $c =0.1$, $k =1.6$. Each column pertains to a different evaluation method (described in \textsection~\ref{sec:eval}). Colors denote the learning method (described in \textsection~\ref{sec:learn}). The observational model performs poorly on the treated population. We see a significant drop in PR and ROC curves between the control evaluation and the true counterfactual evaluation for the observational model. The calibration curves show that the observational model is severely underestimating risk on the treated population. Only the DR and true counterfactual evaluations highlight this significant limitation of the observational model.}
\label{fig:synth_treat_c10_k160}
\end{figure*}

\begin{figure*}
\centering
\begin{subfigure}{\textwidth}
  \centering
  \includegraphics[scale=0.5, trim={0cm 0.5cm 0cm 0cm},clip]{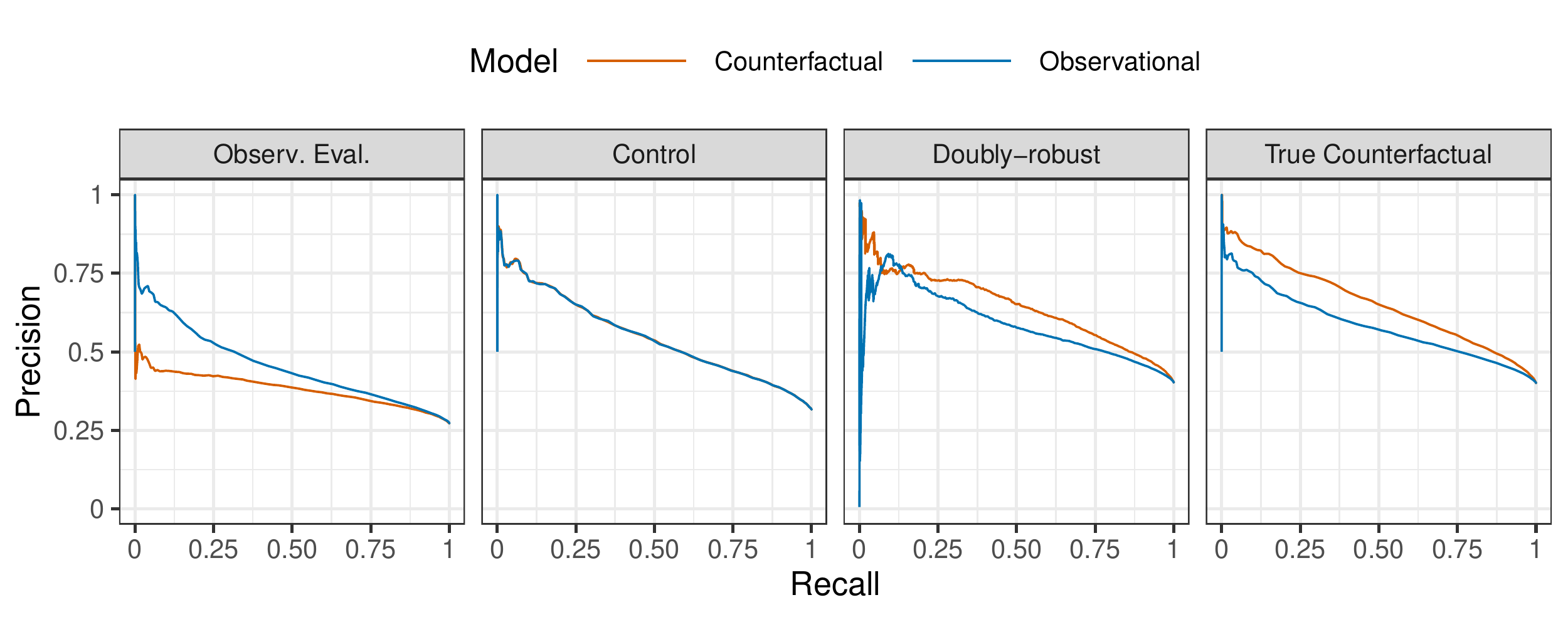}
    \caption{PR curves}
    \label{fig:synth_pr_treat_c50_k160}
\end{subfigure}%
\\
\begin{subfigure}{\textwidth}
  \centering
  \includegraphics[scale=0.5, trim={0cm 0.6cm 0cm 0cm},clip]{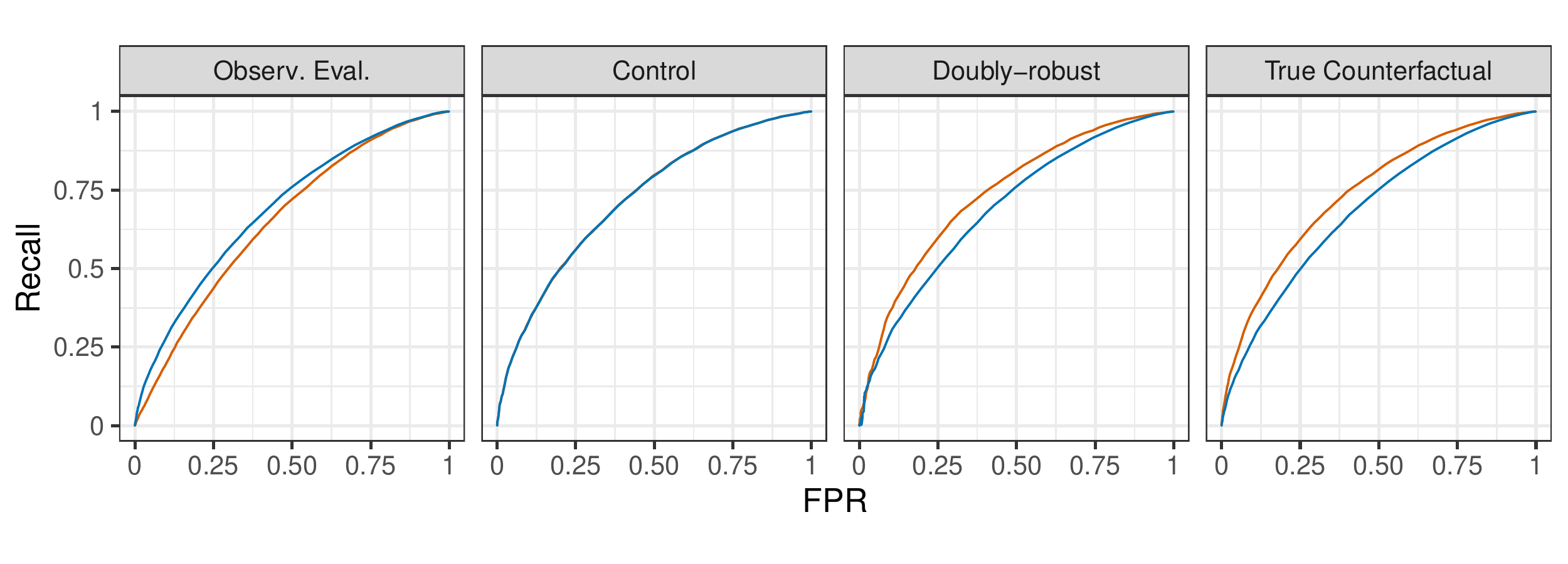}
  \caption{ROC curves}
  \label{fig:synth_roc_treat_c50_k160}
\end{subfigure}
\\
\begin{subfigure}{\textwidth}
  \centering
  \includegraphics[scale=0.5, trim={0cm 0.6cm 0cm 0cm},clip]{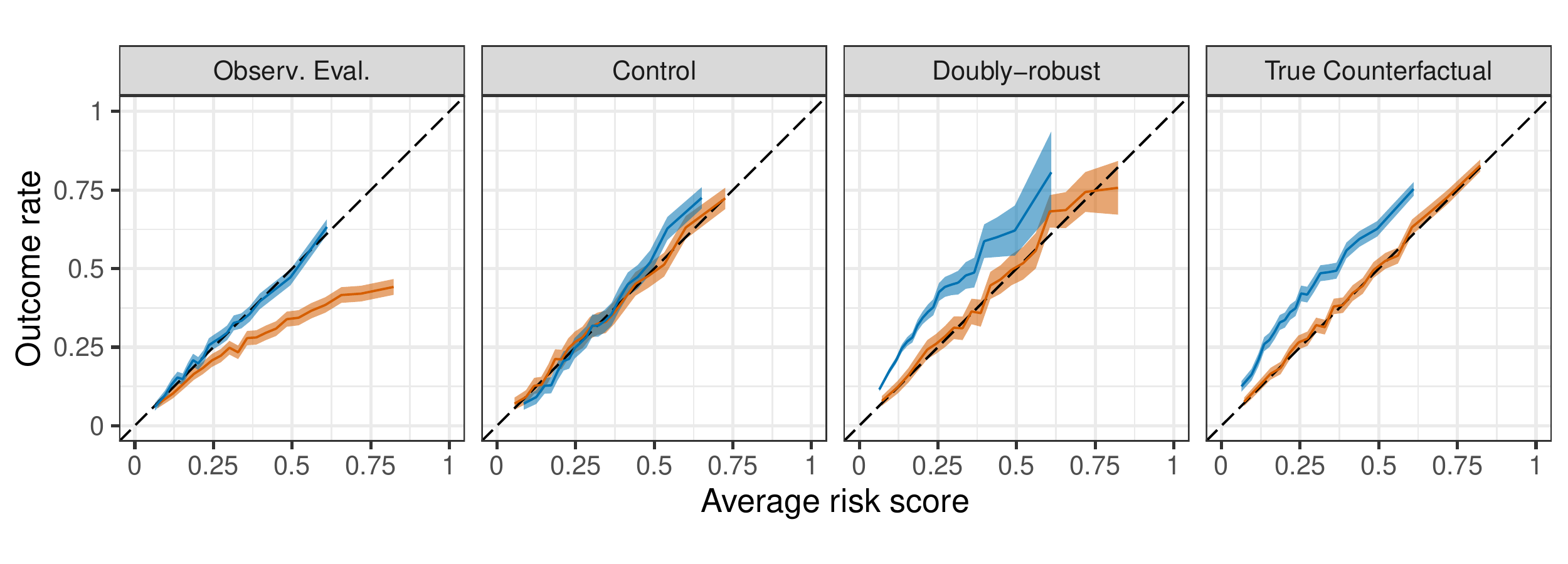}
  \caption{Calibration curves. 95\% pointwise confidence bounds shown.}
  \label{fig:synth_calib_treat_c_k160}
\end{subfigure}

\caption{Synthetic data results when including treatment decision as a feature with parameters with $c =0.5$, $k =1.6$. Each column pertains to a different evaluation method (described in \textsection~\ref{sec:eval}). Colors denote the learning method (described in \textsection~\ref{sec:learn}). Relative to Figure~\ref{fig:synth_treat_c10_k160}, the higher $c$ value here reduces the treatment effect so the observational model performs better on this data, but the observational model still underperforms on the treated population. We see a drop in PR and ROC curves between the control evaluation and the true counterfactual evaluation for the observational model. The calibration curves show that the observational model is  underestimating risk on the treated population.}
\label{fig:synth_treat_c50_k160}
\end{figure*}

In \textsection~\ref{sec:learn}, we presented the observational and counterfactual models of risk and we motivated why the observational model does not appropriately account for treatment effects.
One might wonder if including treatment in the regression could control for treatment effects. 
In Figures~\ref{fig:synth_treat_c10_k160} and~\ref{fig:synth_treat_c50_k160}, we present the PR, ROC, and calibration curves when including $T$ as a feature in the observational model for two values of $c$, the parameter describing treatment effect.
The true counterfactual evaluation shows that the observational model underperforms relative to the counterfactual model. 
This indicates that including $T$ as a feature does not appropriately control for treatment effects for the purposes of building RAIs.

Evaluation based only on the control or observational evaluation methods would lead to the wrong conclusions about model performance.
Comparing the control and true counterfactual evaluations, we see that the observational model severely underperforms on the \emph{treated} population. 
The control evaluation is misleading because it does not account for the poor performance on the treated population; only our DR evaluation (and the true counterfactual evaluation which is not feasible in practice) highlight this significant limitation of the observational model.

\clearpage

\section{Fairness-Corrective Methods on Synthetic Data}\label{sec:appendix_fair_exp}

\vbox
{We supplement the empirical analysis of \textsection~\ref{sec:fair_experiments} by displaying the results as we vary $c$, the parameter describing treatment effect, and $k$, the parameter describing treatment assignment bias. We show the counterfactual ROC curves as well as the table showing the generalized observational and counterfactual FPR and FNR using $t=0.5$ as the threshold.}

\foreach \c/\cd in { 10/.1,  30/.3} 
{ 
\foreach \k/\kd in { 80/.8,  160/1.6} {

\begin{figure*}
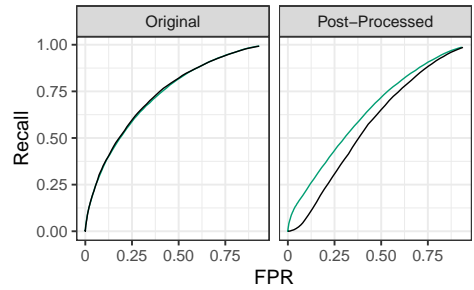

    \centering
    \begin{subfigure}[b]{0.5\textwidth}
    \input{fig/synth_hardt_table_c\c_k\k.tex}
    \vspace*{8mm}
\caption{Observational and counterfactual generalized FPR/FNR for the original and post-processed models}
    \end{subfigure}%
    \begin{subfigure}[b]{0.5\textwidth}
      \includegraphics[scale=.5, trim={0cm 0.7cm 0cm 0cm},clip]{fig/synth_hardt_roc_c\c_k\k.pdf}
      \caption{Counterfactual ROC curves}
    \end{subfigure}
    \caption{Post-processing results on synthetic data with parameters $c =\cd$, $k =\kd$}
\end{figure*}
}
}

\end{document}